\documentclass{article}

\usepackage{PRIMEarxiv}
\usepackage{fancyhdr}       %

\usepackage{microtype}
\usepackage{graphicx}
\usepackage{subcaption}
\usepackage{booktabs} %

\usepackage{hyperref}

\usepackage{algorithmic}

\usepackage{amsmath}
\usepackage{amssymb}
\usepackage{mathtools}
\usepackage{amsthm}

\usepackage[capitalize,noabbrev]{cleveref}

\theoremstyle{plain}
\newtheorem{theorem}{Theorem}[section]
\newtheorem{proposition}[theorem]{Proposition}
\newtheorem{lemma}[theorem]{Lemma}

\theoremstyle{definition}

\newtheorem{assumption}[theorem]{Assumption}
\theoremstyle{remark}

\usepackage{url}
\usepackage{multirow}
\usepackage{sidecap}

\usepackage{natbib}

\usepackage{hyperref}
\usepackage{amsmath,amsfonts,amssymb}
\usepackage{url}
\usepackage{graphicx}
\usepackage{booktabs}
\usepackage{graphicx}      %
\usepackage{wrapfig}       %
\usepackage{multirow}
\usepackage{float}
\usepackage[ruled,vlined,linesnumbered]{algorithm2e}

\begin{document}

\renewcommand{\citep}[1]{\cite{#1}}
\renewcommand{\citet}[1]{Ref~\cite{#1}}

\title{Beyond Accuracy and Complexity: \\ The Effective Information Criterion for Structurally Stable Symbolic Regression}

\author{
  Zihan Yu, Guanren Wang, Jingtao Ding, Huandong Wang, Yong Li \\
  Department of Electronic Engineering, BNRist \\
  Tsinghua University \\
  Beijing, China\\
  \texttt{\{yuzh23, wgr23, dingjt15\}@mails.tsinghua.edu.cn}, \\
  \texttt{\{wanghuandong, liyong07\}@tsinghua.edu.cn}
}

\maketitle

\begin{abstract}
Symbolic regression (SR) traditionally balances accuracy and complexity, implicitly assuming that simpler formulas are structurally more rational. We argue that this assumption is insufficient: existing algorithms often exploit this metric to discover accurate and compact but structurally irrational formulas that are numerically ill-conditioned and physically inexplicable. Inspired by the structural stability of real physical laws, we propose the Effective Information Criterion (EIC) to quantify formula rationality. EIC models formulas as information channels and measures the amplification of inherent rounding noise during recursive calculation, effectively distinguishing physically plausible structures from pathological ones without relying on ground truth. Our analysis reveals a stark structural stability gap between human-derived equations and SR-discovered results. By integrating EIC into SR workflows, we provide explicit structural guidance: for heuristic search, EIC steers algorithms toward stable regions to yield superior Pareto frontiers; for generative models, EIC-based filtering improves pre-training sample efficiency by 2–4 times and boosts generalization $R^2$ by 22.4\%. Finally, an extensive study with 108 human experts shows that EIC aligns with human preferences in 70\% of cases, validating structural stability as a critical prerequisite for human-perceived interpretability. 
\end{abstract}

\section{Introduction}

Symbolic regression (SR) is a machine learning technique that discovers interpretable mathematical formulas describing relationships in data~\citep{camps2023discovering}. Unlike black-box models, it reveals how inputs map to outputs in a form that can be analyzed mathematically or logically, offering insights and new scientific knowledge~\citep{makke2024interpretable}.
In a typical workflow, researchers apply SR algorithms to \textit{data} to generate candidate formulas that balance accuracy and complexity (i.e., the \textit{Pareto front}). They can then assess the credibility of these candidates and select the most reasonable \textit{formula} for insights into the underlying patterns~\citep{liu2024automated}.
Traditionally, SR relies on heuristic search, especially genetic programming, which edits formulas through mutation and crossover to maximize a search objective that balances complexity and accuracy~\citep{cranmer2023interpretable,burlacu2020operon}. Recently, generative methods have gained attention, in which transformer models are pre-trained on synthetic formula--data pairs to predict formula symbols from data.~\citep{biggio2021neural,kamienny2022end,meidani2023snip}. However, trained on synthetic formulas with inevitably irrational structures, these methods typically struggle with sample efficiency and require tens or hundreds of millions of samples to achieve effective generalization in real-world systems~\citep{kamienny2022end,yu2025symbolic}. Recent approaches shift back to the search-based paradigm, integrating pre-trained neural networks~\citep{kamienny2023deep,shojaee2023transformer,yu2025symbolic}, online-trained deep reinforcement learning policies~\citep{petersen2021deep,tenachi2023deep,xu2023rsrm}, or large language models~\citep{shojaeellm} to guide the search algorithms.
 
\begin{figure*}[tb]
    \centering
    \includegraphics[width=\linewidth]{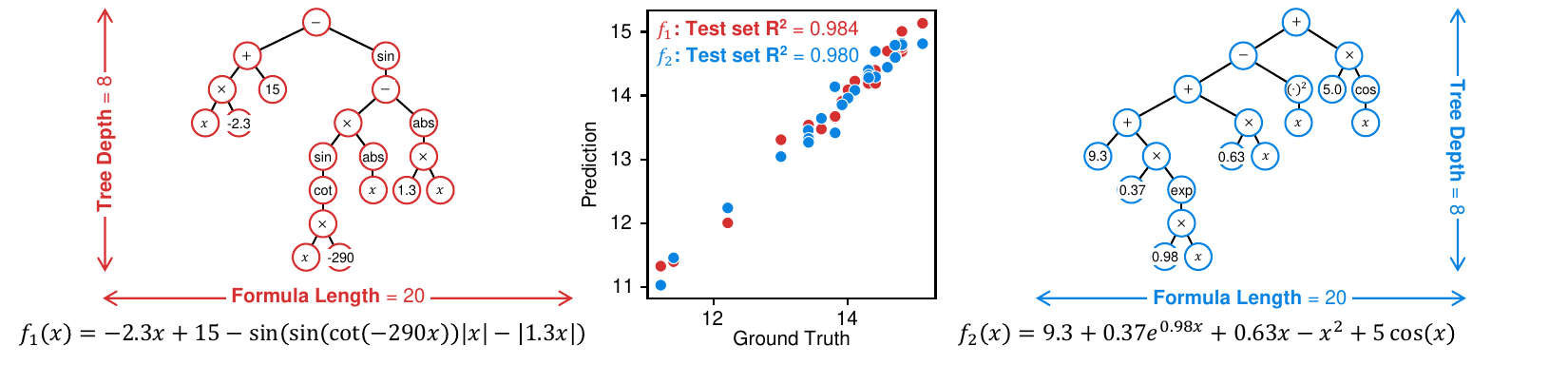}
    \caption{
    \textbf{Formulas with identical complexity and accuracy can exhibit distinct structural rationality.} Despite having the same length and $R^2$, the left formula contains pathological nesting, while the right one remains structurally sound, demonstrating that complexity fails to capture structural rationality.
    }
    \label{fig:fig1}
\end{figure*}

These search-based SR methods typically evaluate candidate formulas based on the trade-off between accuracy and complexity (i.e., the Pareto frontier)~\citep{la2021contemporary}. However, this standard bi-objective framework relies on the implicit assumption that lower complexity implies higher rationality, which is not always valid. The complexity metric, usually measured by the number of symbols in a formula (i.e., formula length)~\citep{la2021contemporary,aldeia2025call}, captures only the overall size but ignores the internal structure of a formula, which is critical for physical plausibility. Driven by powerful search capabilities, existing SR algorithms often discover accurate and compact, but structurally irrational formulas. As illustrated in Figure~\ref{fig:fig1}, although the two formulas possess identical complexity and similar accuracy, they exhibit drastically different structural characteristics: the left one features a nested, irrational structure (e.g., $\sin(\sin(\cot(x)))$) that is physically inexplicable and numerically ill-conditioned, whereas the right one, as a linear combination of simpler functions, aligns well with scientific intuition. The prevalence of such structurally irrational results in practice limits the process of extracting knowledge and hinders the application of SR in scientific discovery.

\begin{figure*}[tb]
    \centering
    \includegraphics[width=\linewidth]{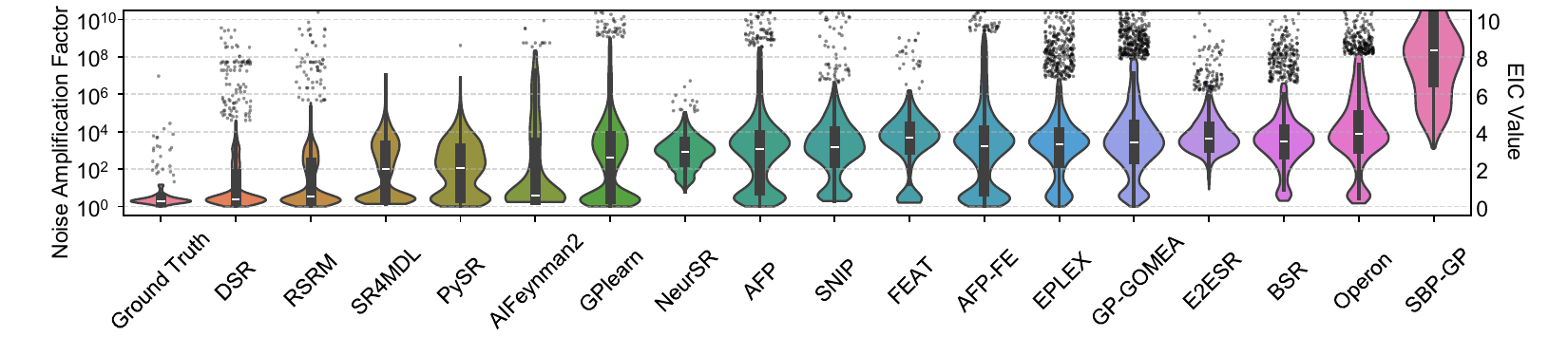}
    \caption{
        \textbf{The structural stability gap between real physical formulas and SR results.}
        We compare 133 ground truth formulas (Feynman, Strogatz) with results from 17 SR algorithms, quantified by the noise amplification factor (Section~\ref{sec:method}). The inner box shows the quartiles, and the black dots indicate outliers beyond $1.5\times$IQR.
    }
    \label{fig:EIC_compare}
\end{figure*}

To address this problem, we draw inspiration from the structural stability of real physical formulas. In Figure~\ref{fig:EIC_compare}, we observe a stark structural stability gap: while human-derived equations are typically robust to computational noise (e.g., floating-point errors), formulas discovered by SR, despite achieving high accuracy and compactness, often suffer from critical structural instability. These mathematically pathological structures drastically amplify rounding noise and, critically, render the formulas irrational and uninterpretable. Consequently, we propose the Effective Information Criterion (EIC) to quantify structural stability. EIC measures the amplification of cumulative noise generated when inherent rounding noise in the calculation propagates in the formula, thereby effectively distinguishing physically plausible formulas from irrational ones.

EIC reveals a stark structural stability gap between real physical formulas and those discovered by existing SR methods. This finding motivates us to integrate EIC into SR workflows to mitigate structural irrationality. Specifically, for heuristic search-based algorithms, EIC guides the search toward stable regions, yielding superior Pareto frontiers while effectively mitigating structural instability. For generative methods, it filters out unstable samples during pre-training, boosting sample efficiency by 2--4 times and enhancing generalization to physical formulas by an $R^{2}$ of 22.4\%. Finally, an evaluation involving 108 human experts confirms that EIC aligns with human preferences in 70\% of cases, validating that structural stability is a critical prerequisite for human-perceived interpretability.

\section{Related Work}

Existing symbolic regression (SR) algorithms can be broadly divided into heuristic search-based and generative methods. Additionally, their metrics and constraints are critical to understanding current limitations.

\noindent\textbf{Heuristic search-based SR methods.} Traditionally, SR relies on heuristic search methods such as genetic programming~\citep{augusto2000symbolic,burlacu2020operon,cranmer2023interpretable} or Monte Carlo tree search~\citep{sun2022symbolic}. Recently, advanced approaches have integrated pre-trained neural networks~\citep{kamienny2023deep,shojaee2023transformer,yu2025symbolic,ying2025neural}, online-trained deep reinforcement learning policies~\citep{petersen2021deep,tenachi2023deep,xu2023rsrm}, or large language models~\citep{shojaeellm} to guide the search algorithms. Furthermore, some works leverage neural networks to identify symmetries in data or perform ab initio variable selection \citep{ye2024ab} to facilitate the search process~\citep{udrescu2020ai}. While these methods typically limit the maximal formula length to avoid excessive complexity, they can barely prevent formulas with irrational structures. This motivates us to incorporate the proposed EIC into the search objective to penalize structurally unstable candidates, thereby improving the rationality of the discovered results.

\noindent\textbf{Generative symbolic regression methods.} In recent years, generative methods have emerged, which pre-train Transformer models on large-scale randomly generated formula -- data pairs to predict mathematical symbols directly from data. These pre-trained models can serve various roles, including generating formulas directly~\citep{biggio2021neural,kamienny2022end}, guiding symbolic search algorithms~\citep{shojaee2023transformer,kamienny2023deep,yu2025symbolic,ying2025neural}, or acting as foundation models for downstream tasks~\citep{meidani2023snip}. Despite their promise, these methods face significant challenges in sample efficiency. The randomly generated training formulas often contain irrational structures that diverge significantly from real-world physical laws, thus requiring extremely large datasets (tens or hundreds of millions of samples) to achieve effective generalization. This motivates us to use EIC to filter out formulas with irrational structures during the pre-training phase, thereby enhancing consistency with real-world formulas and boosting sample efficiency.

\noindent\textbf{Evaluation metrics and constraints used in SR.}
Standard SR benchmarks, such as SRBench~\citep{la2021contemporary}, SRBench++~\citep{de2024srbench++}, SRSD~\citep{matsubara2022srsd}, and LLMSRBench~\citep{shojaee2025llm}, typically employ a bi-objective framework balancing fitting accuracy (e.g., $R^2$, RMSE) and complexity~\citep{fongpareto}. Common complexity metrics like formula length, depth, and minimum description length~\citep{yu2025symbolic} capture overall size but, as Figure~\ref{fig:fig1} illustrates, ignore internal structure.
Moreover, standard evaluations of out-of-distribution (OOD) performance or noise sensitivity focus solely on input-output mappings, overlooking internal rationality. For instance, while $x$ and $(10000x-9999x)$ yield identical mappings, the latter suffers from catastrophic cancellation, amplifying rounding noise by 4 orders of magnitude.
Finally, regarding physical plausibility, methods imposing unit constraints~\citep{tenachi2023deep} are largely limited to physics and difficult to apply in domains like sociology, where units are ambiguous. Meanwhile, manual expert evaluations of trustworthiness either reflect subjective preferences~\citep{lacava20222022} or are too time-consuming for standard benchmarks~\citep{virgolin20232023}.
Consequently, no scalable metric currently evaluates internal structural rationality. EIC fills this gap by explicitly measuring structural stability, complementing existing complexity-based and black-box evaluations.

\section{Effective Information Criterion}
\label{sec:method}
\label{sec:EIC}

\subsection{Intuition and Definition for EIC}
\label{sec:definition}

\noindent\textbf{Motivation}.
As white-box models, symbolic formulas are defined critically by their internal structure, which determines their rationality (Figure~\ref{fig:fig1}), rather than just input-output mappings. We draw inspiration from history: foundational laws were verified using low-precision tools (e.g., slide rules with 3--4 significant digits), implying a strong inductive bias that valid formulas must remain robust under limited precision. Consequently, we propose evaluating rationality via structural stability. Viewing a formula as a computation tree (with inputs as leaves and operators as nodes), intrinsic rounding errors inevitably accumulate and amplify during propagation. Unlike real physical formulas, which are typically stable, SR-discovered models, even those with accurate fitting, often possess ill-conditioned structures that amplify computational noise by orders of magnitude (Figure 2), which renders them irrational. Therefore, we propose quantifying this noise amplification to effectively distinguish real physical formulas from irrational ones.

\noindent\textbf{Mathematical Formalization}. Formally, we represent a formula $f$ as a symbol tree $\mathcal{T}[f]$ (as demonstrated in Figure~\ref{fig:fig1}). Each internal node $k$ implements an elementary operator $e_k$ (e.g., $+$, $\times$, $\sin$) with child nodes $\mathcal{C}[k]$. Given an input sample $x$ drawn from a data distribution $\mathcal{D}$, the ideal, noise-free value $y_k(x)$ at node $k$ can be obtained recursively via $y_k(x) = e_k(\{y_i(x)\}_{i \in \mathcal{C}[k]})$, where $e_k(\cdot)$ represents the ideal operator output computed under infinite precision. Under finite precision, however, the computed value $\tilde{y}_k$ at node $k$ deviates from $y_k$ due to the inherent rounding noise introduced at this node and the propagated error accumulated from the child nodes. Since the magnitude of rounding error is proportional to the value itself, it can be modeled as a multiplicative noise with zero mean and a fixed variance:
\begin{assumption}[Rounding noise as multiplicative noise]
    \label{ass:noise_model}
    \begin{equation}
        \tilde{y}_k = e(\{\tilde{y}_i\}_{i \in \mathcal{C}[k]}) \times (1 + \epsilon_k),
    \end{equation}
    where $\epsilon_k$ is an independent random variable\footnote{Although rounding error is deterministic for a fixed input $x$, quantization theory suggests that as computational precision increases (i.e., variance of the rounding error decreases), the error behaves pseudo-randomly and becomes asymptotically uncorrelated with $x$~\citep{widrow1996statistical}} following any distribution with $\mathbb{E}[\epsilon_k] = 0$ and $\text{Var}[\epsilon_k] = \sigma^2$.
\end{assumption}
To quantify the extent to which the introduced rounding error is amplified due to accumulation during propagation, we define the cumulated error $\eta_k \triangleq (\tilde{y}_k - y_k) / y_k$ and the variance amplification factor
\begin{equation}
s^2_k(x) \triangleq \lim_{\sigma \to 0} \frac{\text{Var}[\eta_k|x]}{\sigma^2},
\end{equation}
which quantifies the sensitivity of the subtree rooted at node $k$ to rounding noise at high precision limits ($\sigma\to 0$) by the degree to which accumulated noise is amplified compared to rounding noise. Note that $s_k(x)$ is a function of $x$ since the same subtree structure can have different sensitivities under different input data. Intuitively, a small $s_k(x)$ suggests the subtree is structurally stable, while a large $s_k(x)$ suggests the subtree acts as a noise amplifier, significantly magnifying small perturbations from upstream calculations.

To analyze how the error accumulation is intrinsically driven by the mathematical structure of the formula, we derive its recursive calculation rule (See Appendix \ref{app:s_k(x)} for detailed derivation):
\begin{proposition}[Recursive Relation of $s_k(x)$]
    \label{prop:s_k(x)}
    \begin{equation}
    \label{eq:recursive_relation}
    s_k^2(x) = 1 + \sum_{i \in \mathcal{C}[k]} \kappa_{k,i}^2(x) \times s_i^2(x),
    \end{equation}
    where $\kappa_{k, i} \triangleq \frac{y_i}{y_k} \frac{\partial e_k}{\partial y_i}$ is the partial relative condition number of the operation $e_k$ with respect to its operand $i$. For leaf nodes where $\mathcal{C}[k] = \emptyset$, we assume its $s_k^2(x) = 1 + 0 = 1$.
\end{proposition}
This reveals that $s_k(x)$ is a deterministic structural property, computable solely from the formula's structure and input data $x$ without relying on external parameters. Physically, it decomposes the total instability into two components, including the \textit{intrinsic rounding noise} generated at the current node (represented by the constant term $1$) and the \textit{propagated noise} from child nodes ($s_i^2(x)$), which is amplified by the squared condition number $\kappa_{k,i}^2$.

\noindent\textbf{Definition of EIC}. Based on the node-wide structural instability measure $s_k(x)$, we define the Effective Information Criterion (EIC) as
\begin{equation}
\label{eq:EIC_def}
\text{EIC} \triangleq \max_{k \in \mathcal{T}[f]} \log_{10} \bar{s}_k.
\end{equation}
Here, $\bar{s}_k^2 \triangleq \mathbb{E}_{x \sim \mathcal{D}} [s_k^2(x)]$ averages local point-wise instability into a global metric. This expectation ensures the evaluation is contextualized within the scientifically relevant domain $\mathcal{D}$, accounting for nonlinear structures where stability fluctuates across regions (e.g., $1/x$ is more unstable when $x$ is smaller). The $\log{10}$ term maps $s_k \in [1, \infty)$ to a linear scale of precision loss, providing clear physical meaning. Crucially, we employ $\max_{k}$ rather than the root metric $s_{\text{root}}$ to impose a strict worst-case constraint. This prevents internal instabilities from being masked by insensitive downstream operations (specifically where $\kappa_{k, i} \approx 0$; see Appendix \ref{app:definition_of_EIC}). This definition, combined with Eq~\eqref{eq:recursive_relation}, allows recursively computing $s_k(x)$ and EIC with $O(N)$ linear complexity via post-order traversal (Appendix Algorithm~\ref{alg:calculate_eic}).

\noindent\textbf{Intuitive understanding of EIC}. 
The EIC quantifies the structural stability of a formula $f$ over the data distribution $\mathcal{D}$ by the amplification of cumulative noise $\eta_k$ generated when inherent rounding noise $\eta_k$ in the calculation propagates in the formula. This can be seen more clearly through (see Appendix~\ref{app:intuitive_of_EIC} for proof)
\begin{equation}
\label{eq:intuitive_of_EIC}
\mathrm{EIC} = \lim_{\sigma \to 0} \max_{k \in \mathcal{T}[f]} \log_{10} \frac{\mathrm{Std}[\eta_k]}{\mathrm{Std}[\epsilon_k]},
\end{equation}
where $\mathrm{Std}[\cdot] = \sqrt{\mathrm{Var}[\cdot]}$ is the standard deviation of random variables that reflects their strength. This relationship clearly illustrates that EIC is the magnitude by which the intensity of the accumulated noise is amplified compared to the added perturbation when a sufficiently small perturbation is added during the calculation of the formula, which is just as shown by the coordinate axes on both sides of Figure~\ref{fig:EIC_compare}.

\noindent\textbf{Theoretical Properties}. 
Besides its intuitive understanding, EIC exhibits several desirable properties.
First, it is parameter-free. As Eq.~\eqref{eq:recursive_relation} shows, calculating $s_k(x)$ depends solely on structure and partial condition numbers, independent of external hyperparameters like noise level $\sigma$.
Second, EIC is scale-invariant. Since condition numbers are dimensionless, EIC remains invariant under unit changes (e.g., rescaling meters to centimeters), as detailed in Appendix~\ref{app:invariance_of_EIC}.
Finally, EIC is sensitive to algebraic rewriting. For instance, $f(x)=x$ and its redundant equivalent $g(x)=\ln(\exp(x))$ yield distinct values. Specifically, when $\mathbb{E}_{\mathcal{D}}[x] \triangleq \mu \gg 1$, we find $\text{EIC}(g) \approx \log_{10}(\mu) > \text{EIC}(f) = 0$ (see Appendix \ref{app:calculation_of_EIC}). Although mathematically equivalent, they differ as calculation graphs: $g$ introduces unnecessary operations that amplify noise. This aligns with parsimony but relies on computational stability rather than symbol counting.

\subsection{Physical Interpretations}
\label{sec:interpretations}

While the definition of EIC clearly elucidates its capability to quantify structural stability, linking it to established physical concepts can further strengthen its theoretical foundation

\noindent\textbf{Numerical Precision Perspective.}
EIC measures structural stability under multiplicative noise, which can be physically interpreted as the loss of significant digits. According to quantization theory~\citep{widrow1996statistical}, the operation of rounding to $N$ significant digits is statistically equivalent (in the variance sense) to injecting multiplicative noise with variance $\sigma^2 \propto 10^{-2N}$. Leveraging this equivalence, we can formally link EIC to the reduction in effective precision (see Appendix~\ref{app:interprelation_for_EIC} for the detailed derivation):
\begin{equation}
\text{EIC} = \lim_{N \to \infty} (N - \min_{k \in \mathcal{T}} M_k),
\end{equation}
where $N$ denotes the working precision in significant digits (corresponding to $\sigma^2$), and $M_k$ represents the effective significant digits remaining at node $k$ after noise accumulation. Therefore, EIC quantifies the maximum magnitude of the intrinsic precision loss imposed by the formula's structure in the asymptotic limit of infinite precision ($N \to \infty$). 
As demonstrated in Figure~\ref{fig:EIC_compare}, EIC values of SR-discovered formulas can reach up to $6$--$8$, implying a catastrophic intrinsic loss of $6$--$8$ significant digits. In sharp contrast, ground-truth physical laws typically exhibit $\text{EIC} < 1$, reflecting a natural preference for structurally robust representations that preserve information fidelity.

\noindent\textbf{Signal Processing Perspective.}
Complementing the static numerical precision view, a dynamic perspective treats the formula as a signal transmission channel, where the input data represents the source signal and the rounding errors act as channel noise. In this signal processing system, noise figure serves as a fundamental metric that quantifies the degradation of the Signal-to-Noise Ratio (SNR) as information traverses the system: $F_k \triangleq 10\log_{10} \frac{\text{SNR}_{\text{intrinsic}}}{\text{SNR}_{k}}$ (in decibels), where $\text{SNR}_{\text{intrinsic}} \triangleq 1/\sigma^2$ represents the theoretical upper bound of signal quality determined by machine precision, and $\text{SNR}_{k} \triangleq 1/\text{Var}[\eta_k]$ denotes the actual SNR remaining at node $k$ after structural noise amplification. As derived in Appendix~\ref{app:interprelation_for_EIC}, EIC can be formally interpreted as the noise figure (scaled by $\frac{1}{20}$) at the bottleneck node:
\begin{equation}
\text{EIC} = \frac{1}{20} \lim_{\sigma \to 0} \max_{k \in \mathcal{T}[f]} F_k.
\end{equation}
This mapping reveals a profound insight: a high EIC indicates that the formula acts as a noisy channel that severely corrupts the information flow. 
For instance, while real physical formulas act as high-fidelity channels (EIC $<1$, or $<20\text{dB}$ loss) that preserve information integrity, a SR-discovered formula with a EIC $\approx 6$ acts as destructive channels with a $120\text{dB}$ SNR drop, implying that the signal quality (SNR) has degraded by twelve orders of magnitude (a $10^{12}\times$ loss in power) relative to the input and rendering the output signal virtually indistinguishable from noise.

\section{Experiments}

In this section, we first reveal the structural stability gap in existing SR methods using EIC (Section~\ref{sec:results1}). Then, we demonstrate the practical utility of EIC in enhancing search-based performance (Section~\ref{sec:EIC_for_search}), improving generative sample efficiency (Section~\ref{sec:EIC_for_generate}), and validating its alignment with human expert preferences for interpretability (Section~\ref{sec:EIC_for_align}).

\subsection{Structural Stability Gap in Existing SR Methods}
\label{sec:results1}

\textbf{Experimental setups.}
To assess the structural rationality of existing SR discoveries, we benchmark 17 symbolic regression methods on 133 white-box problems from SRBench (119 Feynman and 14 Strogatz formulas), comparing the EIC distributions of the discovered formulas against the ground-truth physical formulas. The dataset of each problem is split into 75\% training and 25\% test data. Each method is executed 10 times per problem, with a filtering step that excludes poorly fitting results (Test set $R^2 < 0.8$) to ensure the analysis focuses on accurate candidates.

\begin{figure}[htbp]
    \centering
    \includegraphics[width=0.5\linewidth]{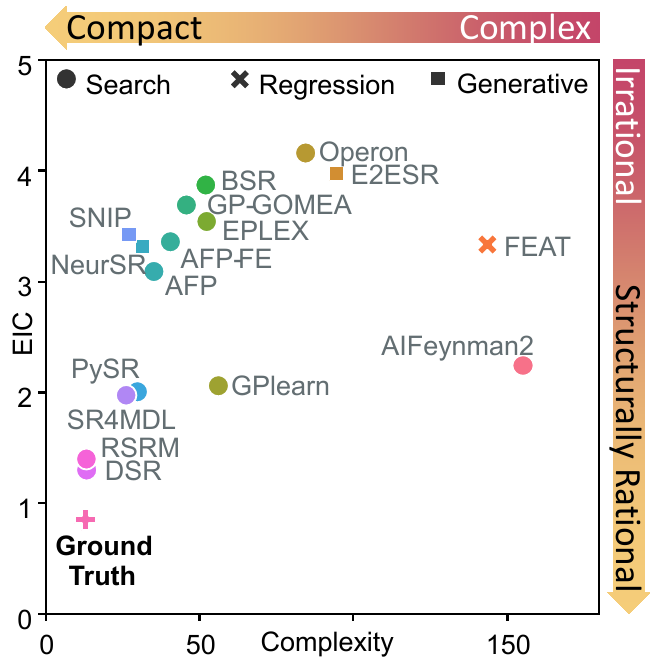}
    \caption{
    \textbf{The structural stability gap between real physical formulas against SR discovered ones.}
    The figure shows the average complexity and EIC of formulas discovered by 17 SR methods on 133 SRBench white-box problems (SBP-GP is not displayed due to its high EIC). The underfit formulas with test set $R^2$ below $0.8$ are ignored to focus on accurate candidates.
    }    
    \label{fig:EIC_vs_Complexity}
\end{figure}

Figure~\ref{fig:EIC_vs_Complexity} presents the average complexity and structural stability (EIC) of formulas discovered by 17 SR methods. Compared to the ground-truth physical formulas, the results reveal the following patterns regarding structural rationality:

\noindent\textbf{Rigid constraints limit expressiveness.}
Methods such as DSR~\citep{petersen2021deep} and RSRM~\citep{xu2023rsrm} achieve the lowest EIC and complexity values, primarily due to their rigorous, manually designed constraints (e.g. forbidding nested trigonometric functions or mutually inverse operators). While these hard constraints effectively prune ill-conditioned structures, they overly restrict the search space, leading to inferior predictive accuracy (as we will demonstrate in the Pareto front analysis in Section~\ref{sec:EIC_for_search}). This highlights the necessity for a flexible, quantitative metric like EIC to guide the search rather than restricting it.

\noindent\textbf{Difference between complexity and structural stability.} Despite exhibiting the highest complexity, AIFeynman2 maintains a moderate EIC. We attribute this to physics-inspired strategies, such as divide-and-conquer and symmetry discovery, which build formulas from physically meaningful components. This confirms EIC identifies AI Feynman's structural rationality, measuring intrinsic robustness rather than merely penalizing length.

\noindent\textbf{Current objectives overlook structural stability.}
Apart from DSR and RSRM that overly restrict the search space, existing approaches like PySR~\citep{cranmer2023interpretable} and SR4MDL~\citep{yu2025symbolic}, despite matching the compactness of the ground truth, fail to match its stability. Specifically, their EIC values ($\approx 2.0$) are significantly higher than real physical formulas ($<1.0$), indicating a difference in noise amplification of over an order of magnitude. This suggests that the standard bi-objective framework (balancing accuracy and complexity, or even absolute Pareto optimality~\citep{fongpareto}) is insufficient to prevent structurally irrational solutions, which motivates us to integrate EIC as a third objective to explicitly steer the search toward structurally rational regimes (as implemented in Section~\ref{sec:EIC_for_search}).

\noindent\textbf{Generative priors bias toward structural instability.} While generative methods like SNIP~\citep{meidani2023snip} and NeurSR~\citep{biggio2021neural} match search-based counterparts (PySR, SR4MDL) in complexity, they exhibit significantly higher EIC. This instability extends to Large Language Models (Appendix~\ref{app:llm_sr}), whose synthesized formulas~\citep{shojaee2025llm} remain far less stable than physical ground truths. This suggests that pre-trained models, despite producing compact formulas, often hallucinate noise-amplifying substructures. We attribute this to the distributional shift between random pre-training data and physical reality, leading to poor generalization. Consequently, we employ EIC to filter pre-training corpora, aligning the training distribution with reality to boost sample efficiency and structural rationality (Section~\ref{sec:EIC_for_generate}).

\subsection{Incorporating EIC into Heuristic Search Methods}
\label{sec:EIC_for_search}

\begin{figure*}[htbp]
    \centering
    \includegraphics[width=0.9\textwidth]{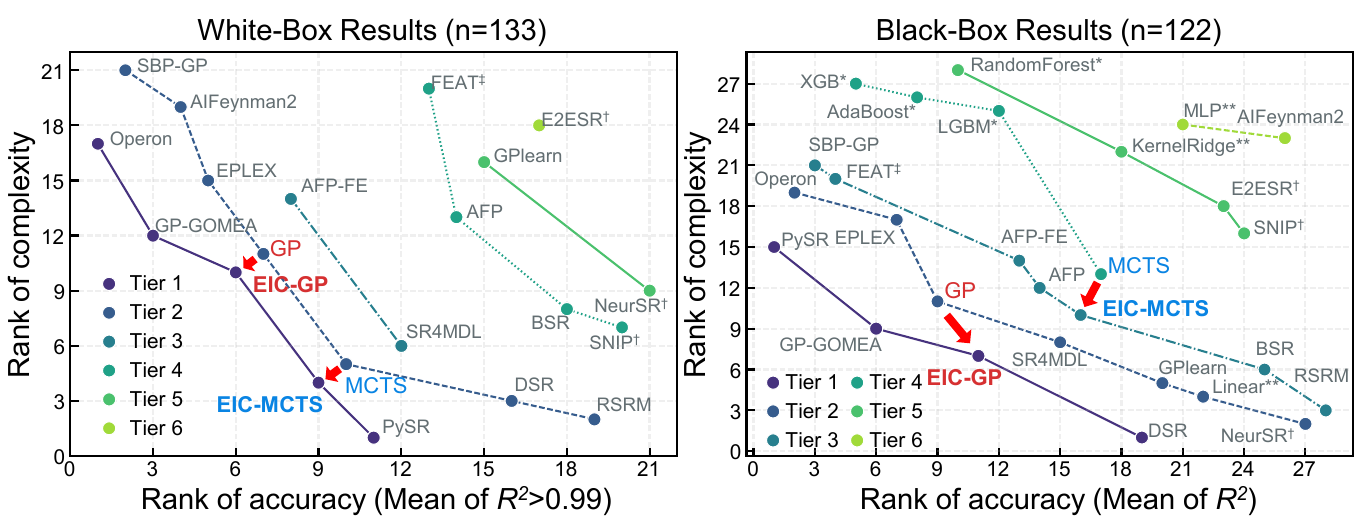}
    \caption{
        \textbf{Pareto fronts on the SR benchmark.}
        $\dagger$, $\ddagger$, $*$, and $**$ indicate generative, regression, decision-tree, and black-box methods, respectively, while others are search-based methods. The lines show the Pareto front tiers, from bottom-left (best) to top-right (worst).
    }
    \label{fig:merged_results}
\end{figure*}

\textbf{Experimental setups.}
To evaluate EIC's efficacy in guiding search, we integrated it into two representative heuristic algorithms: Genetic Programming (GP)~\citep{augusto2000symbolic} and Monte Carlo Tree Search (MCTS)~\citep{sun2022symbolic}, which underpin widespread methods like PySR, TPSR~\citep{shojaee2023transformer}, SR4MDL, etc. We modify the standard fitness function $f = \eta^{C}/(1+\text{NMSE})$, where $C$ denotes the formula complexity (length), $\text{NMSE}=\text{MSE}/\text{Var}(y)$, $\eta < 1$ penalizes formula complexity~\citep{sun2022symbolic}, introducing an EIC penalty term: $f_{\text{new}} = f - \alpha \cdot \text{EIC}$, where $\alpha$ steers the search away from structurally irrational solutions (see Appendix~\ref{app:eic_for_search} for details). Notably, we employ linear regression to optimize constants during the search, avoiding more costly nonlinear optimization algorithms such as BFGS\citep{fletcher2000practical} and accelerating the search. We benchmark these EIC-enhanced variants against 17 baselines on SRBench, comprising 133 white-box and 122 black-box problems. For each problem, we conduct 10 independent trials with a 4-hour limit using a 75\%/25\% train-test split. For white-box tasks, we further assess robustness by injecting Gaussian noise at three levels $\{10^{-3}, 10^{-2}, 10^{-1}\}$. Performance is evaluated based on the Accuracy-Complexity Pareto frontier, which is the standard metric in SR. Besides, we also evaluate the results' EIC to assess structural rationality.

\noindent\textbf{White-box experimental results.}
Figure~\ref{fig:merged_results} (left) summarizes white-box results (full results in Appendix~\ref{app:search_results}). Integrating EIC advanced MCTS and GP from the second to the first Pareto tier, improving both their accuracy and compactness. This improvement can be attributed to EIC explicitly penalizing structurally irrational patterns, which prevents excessive operator stacking and acts as a regularizer to enhance generalization. Consequently, although EIC targets an objective distinct from the standard Accuracy-Complexity trade-off, it synergistically improves both metrics. This structural improvement is corroborated by Appendix Figure~\ref{fig:app_eic_dist}, where EIC values for both methods drop significantly toward ground-truth levels—a natural consequence of explicitly minimizing EIC during the search. In contrast, while DSR and RSRM yield compact formulas with low EIC (Figure~\ref{fig:EIC_vs_Complexity}), their significantly lower accuracy suggests that their manual constraints, while effectively avoiding unreasonable structures, inevitably shrink the search space and limit expressiveness.

\noindent\textbf{Black-box experimental results.}
Figure~\ref{fig:merged_results} (right) presents results on 122 black-box problems (full results in Appendix~\ref{app:search_results}). Incorporating EIC advanced the Pareto fronts of GP (Tier 2 $\to$ 1) and MCTS (Tier 4 $\to$ 3), demonstrating that excluding structurally unreasonable formulas enhances performance even without ground truth. This is corroborated by significantly reduced EIC values in Appendix Figure~\ref{fig:app_eic_dist}. Notably, this reduction is more pronounced than in white-box settings, suggesting EIC effectively counters SR's tendency to rely on overly complex forms when underlying relationships are opaque. However, while MCTS improved in both accuracy and simplicity, GP improved simplicity but slightly reduced accuracy. This is probably due to a trade-off between numerical approximation and structural rationality. Standard GP often exploits numerically unstable structures (e.g., high-degree polynomials) as universal approximators to minimize error, yielding accurate but physically meaningless formulas. EIC constrains the search to stable forms, purposely sacrificing the marginal accuracy gained from such unstable approximations to ensure physical plausibility.

\subsection{Incorporate EIC into Generative SR Methods}
\label{sec:EIC_for_generate}

Current generative approaches, such as E2ESR~\citep{kamienny2022end}, SNIP~\citep{meidani2023snip}, and SR4MDL~\citep{yu2025symbolic}, typically rely on the random formula generation algorithm developed by \citet{lampledeep} for pretraining. While this mechanism provides unlimited training data, the generated synthetic formulas often exhibit unstable structures rarely seen in real-world physical equations. Consequently, models pre-trained on such data suffer from severe sample inefficiency, often requiring tens of millions of samples to achieve effective generalization.

To address this, we propose an EIC-guided rejection sampling strategy to filter the pretraining corpus. Specifically, during data generation, we evaluate the EIC value of each sample and discard those exceeding a threshold $\theta$, regenerating them until the condition is met. This process removes structurally implausible samples while preserving the overall diversity of the training data. Based on the distribution of real physical formulas shown in Figure 2, we set $\theta=2$ to enforce structural alignment with physical reality (see Appendix~\ref{app:eic_for_generative} for details).

\textbf{Experimental setups.}
In this experiment, we applied this EIC-guided rejection sampling strategy to E2ESR, SNIP, and SR4MDL, covering the progression from early baselines to state-of-the-art methods. Each method was trained on both unfiltered (random) and filtered ($\text{EIC} < 2$) formulas, and evaluated on the 133 SRBench white-box problems. Performance was measured using $R^2$ for E2ESR and SNIP, and RMSE/MAE for SR4MDL\footnote{Strictly speaking, SR4MDL predicts minimum description length rather than tokens. However, as it shares the same architecture and data scheme as E2ESR, we include it for comparison.}. Training on unfiltered data was conducted until test performance plateaued (approx. 5 million samples), serving as the baseline target to evaluate the sample efficiency gains of the EIC-filtered training.

Additionally, we benchmark against a recent data construction baseline, PhyE2E~\citep{ying2025neural}, which uses LLMs to generate ``look-physical'' formulas with unit constraints. We trained E2ESR on 180k PhyE2E formulas. To ensure a fair comparison, given its limited dataset size, we employed a mixed sampling strategy: sampling from PhyE2E formulas with a probability of 0.1 and generating random formulas with a probability of 0.9.

\noindent\textbf{EIC-guided filtering reduces the distribution gap between training samples and real physical formulas.}
We evaluated whether EIC filtering produces a training distribution closer to real-world formulas. We generated 1024 formulas from the unfiltered random generator and 1024 from the filtered generator, and compared them with three benchmark sets, including Feynman (119 physics formulas), Strogatz (14 ODE formulas), and Wiki Named Equations (984 equations, see Appendix~\ref{app:generative_results} for details). Similarity was measured via variable counts, constant counts, operator counts, and formula length, using Jensen-Shannon (JS) and Kullback-Leibler (KL) divergences.

Table~\ref{tab:distribution_difference} shows that EIC filtering yields distributions substantially closer to real formulas, with $20\sim50\%$ reduction in JS divergence and $30\sim70\%$ reduction in KL divergence compared to the unfiltered random formulas. This demonstrates that EIC effectively reduces the gap between synthetic and real-world formulas and improves train-test alignment.

\begin{figure}[htbp]
    \centering
    \includegraphics[width=0.3\linewidth]{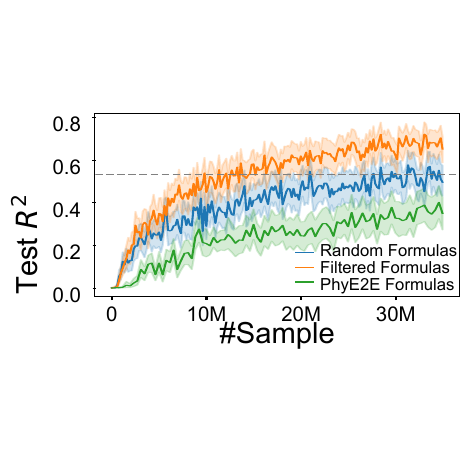}
    \caption{
    \textbf{Generalization performance of E2ESR trained on different samples.} 
    The grey line shows the convergence trained on random formulas.
    }
    \label{fig:improve_E2ESR}
\end{figure}

\noindent\textbf{EIC-guided filtering boosts sample efficiency and generalization performance.}
The results are summarized in Table~\ref{tab:sample_efficiency_improvement}, where models trained on randomly generated formulas eventually converged on the Feynman test set, but typically required tens of millions of samples. In contrast, training on EIC-filtered formulas achieved equal or better performance with fewer training samples, improving sample efficiency by 357\%, 233\%, and 287\% on E2ESR, SNIP, and SR4MDL, respectively. This demonstrates that EIC filtering removes structurally unreasonable formulas, producing training data that is more physically meaningful and transferable to real-world tasks, thereby enhancing pretraining-based symbolic regression.

Figure~\ref{fig:improve_E2ESR} shows that training on EIC-filtered samples further improved final performance compared to random formulas. For E2ESR, training for the same number of steps increased $R^2$ from $0.55$ to $0.68$, representing a 22.4\% relative gain. Similar improvements were observed for SNIP and SR4MDL, with relative gains of 13.5\% and 5.14\%, respectively (see Appendix~\ref{app:generative_results}). Incorporating PhyE2E formulas into pretraining did not improve performance and slightly reduced generalization, likely due to a trade-off between unit consistency and formula diversity: although PhyE2E formulas satisfy unit constraints, their limited diversity reduces pretraining effectiveness. In contrast, EIC filtering preserves both structural plausibility and distributional diversity, thus yielding superior generalization.

\begin{table}[tb]
    \centering
    \caption{
    \textbf{Distribution differences between real physical formulas and generated ones.} The JS and KL divergence between three real physical formula sets: \textbf{F}eynman ($n=119$), \textbf{S}trogatz ($n=14$), and \textbf{W}iki Named Equations datasets ($n=984$), as well as 1024 random formulas and 1024 EIC-filtered ones ($\mathrm{EIC}<2.0$).
    }
    \label{tab:distribution_difference}

\begin{tabular}{rlccccccccccc}
\toprule
\multicolumn{1}{l}{\textbf{}} &   & \multicolumn{2}{c}{\textbf{\#Variables}} &   & \multicolumn{2}{c}{\textbf{\#Coefficients}} &   & \multicolumn{2}{c}{\textbf{\#Operators}} &   & \multicolumn{2}{c}{\textbf{Formula Length}} \\
\cmidrule{3-4}\cmidrule{6-7}\cmidrule{9-10}\cmidrule{12-13}  &   & \textbf{JS↓} & \textbf{KL↓} &   & \textbf{JS↓} & \textbf{KL↓} &   & \textbf{JS↓} & \textbf{KL↓} &   & \textbf{JS↓} & \textbf{KL↓} \\
\midrule
\multicolumn{1}{l}{\textbf{F.}} & Naive & 0.1196 & 1.4390 &   & 0.5163 & 16.78 &   & 0.3233 & 9.539 &   & 0.3973 & 13.615 \\
 & Filtered & 0.0768 & 0.3523 &   & 0.3769 & 5.582 &   & 0.1862 & 0.7741 &   & 0.2573 & 3.4397 \\
  & $-\Delta$(\%) & \textbf{36\%} & \textbf{76\%} &   & \textbf{27\%} & \textbf{67\%} &   & \textbf{42\%} & \textbf{92\%} &   & \textbf{35\%} & \textbf{75\%} \\
\midrule
\multicolumn{1}{l}{\textbf{S.}} & Naive & 0.4014 & 18.784 &   & 0.6016 & 23.37 &   & 0.5726 & 22.24 &   & 0.6002 & 22.27 \\
 & Filtered & 0.2650 & 12.920 &   & 0.5048 & 20.78 &   & 0.4646 & 18.74 &   & 0.5225 & 20.05 \\
  & $-\Delta$(\%) & \textbf{34\%} & \textbf{31\%} &   & \textbf{16\%} & \textbf{11\%} &   & \textbf{19\%} & \textbf{16\%} &   & \textbf{13\%} & \textbf{10\%} \\
\midrule
\multicolumn{1}{l}{\textbf{W.}} & Naive & 0.1459 & 0.6654 &   & 0.5566 & 14.888 &   & 0.3873 & 4.269 &   & 0.4219 & 7.060 \\
 & Filtered & 0.0747 & 0.2899 &   & 0.4660 & 3.5307 &   & 0.2485 & 1.124 &   & 0.3114 & 1.541 \\
  & $-\Delta$(\%) & \textbf{49\%} & \textbf{56\%} &   & \textbf{16\%} & \textbf{76\%} &   & \textbf{36\%} & \textbf{74\%} &   & \textbf{26\%} & \textbf{78\%} \\
\bottomrule
\end{tabular}%

\end{table}

\begin{table}[tb]
    \centering
    \caption{
        \textbf{Sample efficiency of different methods trained with random and filtered formulas.}
        \#S is the number of samples.
    }
    \label{tab:sample_efficiency_improvement}
        
\begin{tabular}{lccccccccc}
\toprule
\textbf{Pre-training} & \multicolumn{2}{c}{\textbf{E2ESR}} &   & \multicolumn{2}{c}{\textbf{SNIP}} &   & \multicolumn{3}{c}{\textbf{SR4MDL}} \\
\cmidrule{2-3}\cmidrule{5-6}\cmidrule{8-10}
\textbf{Formulas} & \#S↓ & $R^2$↑ &   & \#S↓ & $R^2$↑ &   & \#S↓ & MAE↓ & RMSE↓ \\
\midrule
Random & 35M & 0.5190 &   & 40M & 0.5300 &   & 50M & 6.972 & 8.731 \\
Filtered & \textbf{9.79M} & \textbf{0.5399} &   & \textbf{25.15M} & \textbf{0.5415} &   & \textbf{17.4M} & \textbf{6.812} & \textbf{8.701} \\
\midrule
$\Delta\text{Efficiency}$ & \multicolumn{2}{c}{357.5\%} &   & \multicolumn{2}{c}{198.8\%} &   & \multicolumn{3}{c}{287.4\%} \\
\bottomrule
\end{tabular}%

\end{table}

\subsection{EIC Aligns with Human Preferences for Interpretability}
\label{sec:EIC_for_align}

\noindent\textbf{Experimental setups.}
To validate EIC's alignment with human interpretability preferences, we conducted a controlled study comparing formula pairs with similar accuracy and complexity but distinct EIC values. We pooled formulas discovered by search-based symbolic regression methods across black-box problems (Section \ref{sec:EIC_for_search}). To ensure intuitive assessment, we restricted analysis to 1D and 2D problems allowing clear visualization. From this pool, we rigorously selected pairs with indistinguishable fitting accuracy ($\Delta R^2 \le 0.02$) and complexity ($\Delta C \le 2$), yet significant structural divergence ($|\Delta \text{EIC}| \ge 2$). We invited 108 science/engineering volunteers to evaluate 10 random pairs each, selecting the more interpretable formula based on visualized behavior and mathematical form. In total, 870 valid evaluations were collected (see Appendix \ref{app:expert_rating}).

\begin{figure}[htbp]
    \centering
    \includegraphics[width=\linewidth]{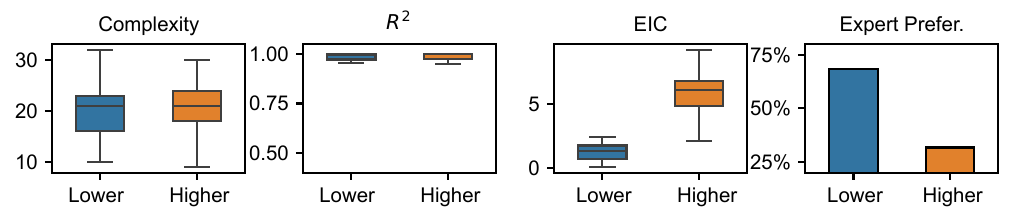}
    \caption{
        \textbf{Comparison of formula pairs}. ``Lower'' and ``Higher'' are formulas with the relatively lower/higher EIC values within each pair, whose complexity and accuracy ($R^2$) are indistinguishable but elicit a decisive expert preference for the lower candidates.
    }
    \label{fig:LLM_rating}
\end{figure}

\noindent\textbf{EIC Aligns with Human Expert Preferences for Interpretability.}
The results, summarized in Figure~\ref{fig:LLM_rating}, demonstrate a strong concordance between EIC scores and expert preferences. 
Although the paired formulas possess comparable distributions of complexity and accuracy (left panels), human experts exhibited a decisive preference for the formulas with lower EIC scores, favoring them in 69.8\% of evaluations (95\% CI: [67\%, 73\%]). 
A one-sided binomial test confirms this preference is statistically significant far beyond random chance ($p<10^{-30}$). 
This result indicates that when accuracy and complexity are controlled, EIC serves as a critical third dimension that effectively captures the structural rationality preferred by humans. 
To assess the reliability of these subjective judgments, we calculated the inter-rater reliability using Fleiss' Kappa, yielding $\kappa=0.37$. 
This indicates a fair level of agreement among experts, suggesting that EIC captures an objective structural quality recognized by humans rather than random individual preferences. 
Furthermore, Large Language Models (LLMs) acting as domain experts exhibited a similar preference pattern (favoring low-EIC formulas in 72.19\% of cases), providing independent corroboration of the human rating results.

\section{Conclusion and Discussion}

We propose the Effective Information Criterion (EIC), a metric to quantify the structural stability of symbolic regression formulas. EIC measures the amplification of inherent rounding noise during recursive calculation, offering a grounded interpretation from both numerical precision and signal-processing perspectives. Unlike traditional complexity-based metrics, EIC effectively evaluates structural rationality and detects ill-conditioned structures that render formulas physically implausible. Our results demonstrate that integrating EIC can steer search-based SR methods away from irrational formulas to yield superior Pareto frontiers and can boost the sample efficiency of generative SR methods by filtering implausible samples. Furthermore, EIC aligns with human expert preference for interpretability in 70\% of cases, validating structural stability as a critical prerequisite for human-perceived interpretability.
Despite its effectiveness, several avenues remain for future work. First, integrating EIC guidance with pre-trained model guidance may further boost performance. Second, analyzing EIC at each subformula, rather than only the maximum, may more precisely identify redundant or unreasonable structures, enabling finer-grained evaluation. Finally, in complex systems where simple microscopic rules produce emergent macroscopic behaviors, such as phase transitions, it remains to be studied whether the resulting dynamics formulas retain low EIC scores.

\clearpage

\section*{Impact Statement}
This paper presents the Effective Information Criterion (EIC), a metric designed to quantify the structural rationality of formulas discovered by symbolic regression algorithms. By prioritizing formulas with structural stability, our work contributes to more reliable AI-driven scientific discovery. We do not foresee any direct negative social impacts from this research.

\bibliographystyle{unsrt}
\bibliography{reference}

\appendix

\section{Effective Information Criterion (EIC)}

\subsection{Proof of Proposition \ref{prop:s_k(x)}}
\label{app:s_k(x)}

\begin{proof}

First, we show that the cumulated error $\eta_k$ is a zero-mean random variable with a variance that tends to 0 as $\sigma \to 0$.

\begin{lemma}
    \label{prop:recursive_error}
    Under a first-order approximation, $\eta_k$ is a zero-mean random variable given by the recursive linearization:
    \begin{equation}
        \eta_k \approx \epsilon_k + \sum_{i\in\mathcal{C}[k]} \kappa_{k, i} \times \eta_i,
        \label{eq:linearized_error}
    \end{equation}
    where $\kappa_{k, i} \triangleq \frac{y_i}{y_k} \frac{\partial e_k}{\partial y_i}$ is the partial relative condition number of the operation at node $k$ with respect to input $i$. The approximation becomes exact in the limit $\sigma \to 0$.
\begin{proof}
    Let the inputs to node $k$ be $y_1, \dots, y_i, \dots, y_m$, which correspond to the outputs of child nodes. For the leaf nodes, we have $\tilde{y}_k = y_k (1 + \epsilon_k)$ and $\eta_k = (\tilde{y}_k - y_k) / y_k$, so $\eta_k = \epsilon_k$ for leafe nodes, whose mean is zero and whose variance, $\sigma$, tends to zero when $\sigma \to 0$. Then, according to Assumption \ref{ass:noise_model}, we have
    \begin{equation}
        \tilde{y}_k = e_k(\tilde{y}_1, \dots, \tilde{y}_m) \times (1 + \epsilon_k)
    \end{equation}
    Substituting $\tilde{y}_i$:
    \begin{equation}
        \tilde{y}_k = e_k(y_1 \times (1+\eta_1), \dots, y_m \times (1+\eta_m)) \times (1 + \epsilon_k)
    \end{equation}
    
    We perform a first-order Taylor expansion of $e$ around the noise-free values $\mathbf{y} = (y_1, \dots, y_m)$, obtaining
    \begin{equation}
        e_k(y_1 \times (1+\eta_1), \dots, y_m \times (1+\eta_m)) \approx e_k(\mathbf{y}) + \sum_{i=1}^m \frac{\partial e_k}{\partial y_i} (y_i \eta_i)
    \end{equation}
    Substituting this back into the expression for $\tilde{y}_k$, and noting that $y_k = e_k(\mathbf{y})$:
    \begin{equation}
        \tilde{y}_k \approx \left( y_k + \sum_{i=1}^m \frac{\partial e_k}{\partial y_i} y_i \eta_i \right) \times (1 + \epsilon_k)
    \end{equation}
    Expanding the product and ignoring the second-order error term ($\eta_i \cdot \epsilon_k \ll \min\{\eta_i, \epsilon_k\}$, since $|\eta_i| \ll 1 ~\text{and}~ |\epsilon_k| \ll 1 ~\text{when}~ \sigma \rightarrow 0$):
    \begin{equation}
        \tilde{y}_k \approx y_k + y_k \epsilon_k + \sum_{i=1}^m \frac{\partial e_k}{\partial y_i} y_i \eta_i
    \end{equation}
    Dividing both sides by $y_k$, we identify the accumulated relative error $\eta_k \triangleq (\tilde{y}_k - y_k)/y_k$ as:
    \begin{equation}
        \eta_k \approx \epsilon_k + \sum_{i=1}^m \underbrace{\left( \frac{y_i}{y_k} \frac{\partial e_k}{\partial y_i} \right)}_{\kappa_{k,i}} \eta_i
    \end{equation}
    
    Since $\epsilon_k$ has zero mean and is independent of $\eta_i$ (which are linear combinations of zero-mean noise from previous steps and is thus independent of each other), the expectation $\mathbb{E}[\eta_k]$ is a linear combination of zeros, satisfying $\mathbb{E}[\eta_k] = 0$, and its variance $\mathrm{Var}[\eta_k] \rightarrow 0$ when $\sigma \rightarrow 0$ (since both $\epsilon_k$ and $\eta_i$ $\rightarrow 0$), which further suggests $|\eta_k| \rightarrow 0$ when $\sigma \rightarrow 0$. This confirms that the cumulated error $\eta_k$ is a zero-mean random variable whose variance tends to zero when $\sigma \to 0$.
\end{proof}
\end{lemma}

Then, we can prove the Eq~\eqref{eq:recursive_relation} by applying the conditional variance operator $\text{Var}[\cdot|x]$ to both sides in Eq~\eqref{eq:linearized_error}. Two key independence properties apply here:
\begin{enumerate}
    \item $\epsilon_k$ is the intrinsic noise introduced at the current node and is independent of the propagated errors $\{\eta_i\}_{i \in \mathcal{C}[k]}$ from previous steps.
    \item Since the formula is structured as a computational tree, the sub-expressions corresponding to different children are disjoint. Thus, the accumulated errors $\eta_i$ from different children are mutually independent random variables.
\end{enumerate}
Using these independence properties, the variance of the sum becomes the sum of the variances:
\begin{equation}
    \text{Var}[\eta_k|x] \approx \text{Var}[\epsilon_k] + \sum_{i \in \mathcal{C}[k]} \kappa_{k,i}^2(x) \text{Var}[\eta_i|x],
\end{equation}
where the approximation becomes exact in the limit $\sigma \to 0$.

Substituting $\text{Var}[\epsilon_k] = \sigma^2$, dividing both sides by $\sigma^2$, and taking the limit $\sigma \to 0$ (which makes the linear approximation exact), we get
\begin{equation}
    \lim_{\sigma \to 0} \frac{\text{Var}[\eta_k | x]}{\sigma^2} = 1 + \sum_{i \in \mathcal{C}[k]} \kappa_{k,i}^2(x) \left( \lim_{\sigma \to 0} \frac{\text{Var}[\eta_i | x]}{\sigma^2} \right)
\end{equation}
which provides the recursive relation to conclude the proof:
\begin{equation}
    s_k^2(x) = 1 + \sum_{i \in \mathcal{C}[k]} \kappa_{k,i}^2(x) s_i^2(x)
\end{equation}

\end{proof}

\subsection{Remark for the definition of EIC}
\label{app:definition_of_EIC}
In Eq~\eqref{eq:EIC_def}, we define EIC as the maximum value of the logarithm $\bar{s}_k$ of all nodes $k \in \mathcal{T}[f]$, rather than the logarithm of the root node $\bar{s}_\text{root}$. The choice of the maximum operator is motivated by the non-monotonic nature of error propagation. Observing the recurrence relation in eq~\eqref{eq:recursive_relation}, if the squared relative condition number is small ($\kappa_{k,i}^2 \ll 1$), the cumulative relative error $s_k(x)$ does not necessarily increase relative to its children's $S_i$. Physically, a vanishing $\kappa_{k,i}$ implies that the child node $i$ exerts negligible influence on its parent $k$. In symbolic regression, this indicates \textit{structural redundancy}, that is, the existence of components that do not contribute to describing the data, which violates the principle of Occam's Razor~\citep{domingos1999role}. Crucially, such a redundant sub-structure $i$ can be numerically ill-conditioned (exhibiting a very high $s_i(x)$), yet its instability is masked by the small $\kappa_{k,i}$, resulting in a deceptively low $s_{\text{root}}(x)$. By defining EIC as the maximum over all nodes ($\max_{k}$), we explicitly detect and penalize these \textit{redundant yet ill-conditioned} internal structures, ensuring that a physically plausible formula implies stability not just at the output, but throughout its entire computational graph.

\subsection{Proof of Eq~\eqref{eq:intuitive_of_EIC}}
\label{app:intuitive_of_EIC}
\begin{proof}
    Based on the definition of EIC and $\bar{s}_k, s_k(x)$, we have
    \begin{equation}
    \begin{aligned}
    \text{EIC} 
    &\triangleq \max_{k \in \mathcal{T}[f]} \log_{10} \bar{s}_k \\
    &\triangleq \max_{k \in \mathcal{T}[f]} \log_{10} \sqrt{\mathbb{E}_{x\sim\mathcal{D}}[s_k^2(x)]} \\
    &= \frac{1}{2} \max_{k \in \mathcal{T}[f]} \log_{10} \mathbb{E}_{x\sim\mathcal{D}}[s_k^2(x)] \\
    &\triangleq \frac{1}{2} \max_{k \in \mathcal{T}[f]} \log_{10} \mathbb{E}_{x\sim\mathcal{D}} \left[ \lim_{\sigma \to 0} \frac{\text{Var}[\eta_k | x]}{\sigma^2} \right]
    \end{aligned}
    \end{equation}

    Here, the limit operator can be sequentially commuted with the expectation, logarithm, and maximum operators due to the linearity of expectation, the continuity of the logarithm, and the finiteness of the node set $\mathcal{T}[f]$, respectively:
    \begin{equation}
    \begin{aligned}
    \label{eq:tmp}
    \text{EIC} 
    &= \frac{1}{2} \max_{k \in \mathcal{T}[f]} \log_{10} \lim_{\sigma \to 0} \mathbb{E}_{x\sim\mathcal{D}} \left[ \frac{\text{Var}[\eta_k | x]}{\sigma^2} \right] \\
    &= \frac{1}{2} \max_{k \in \mathcal{T}[f]} \lim_{\sigma \to 0} \log_{10} \mathbb{E}_{x\sim\mathcal{D}} \left[ \frac{\text{Var}[\eta_k | x]}{\sigma^2} \right] \\
    &= \frac{1}{2} \lim_{\sigma \to 0} \max_{k \in \mathcal{T}[f]} \log_{10} \mathbb{E}_{x\sim\mathcal{D}} \left[ \frac{\text{Var}[\eta_k | x]}{\sigma^2} \right]
    \end{aligned}
    \end{equation}

    Finally, to merge the conditional variance and expectation into the global variance, consider law of total variance decomposition:
    \begin{equation}
        \mathbb{E}_x [\text{Var}[\eta_k | x]] = \text{Var}[\eta_k] - \text{Var}_x [\mathbb{E}[\eta_k | x]],
    \end{equation}
    where the second term is a higher-order infinitesimal of the first term when $\sigma \to 0$: consider that the conditional mean relative error is $\mathbb{E}[\eta_k | x] \approx 0$. The non-zero bias arises only from second-order terms (e.g., $\eta^2$), implying $\mathbb{E}[\eta_k | x] = O(\sigma^2)$. Consequently, the variance of this bias scales as the square of the order:
    \begin{equation}
        \text{Var}_x [\mathbb{E}[\eta_k | x]] = \text{Var}[O(\sigma^2)] = O(\sigma^4).
    \end{equation}
    Substituting this back, the fraction inside the limit becomes:
    \begin{equation}
        \frac{\mathbb{E}_x [\text{Var}[\eta_k | x]]}{\sigma^2} = \frac{\text{Var}[\eta_k] - O(\sigma^4)}{\sigma^2} = \frac{\text{Var}[\eta_k]}{\sigma^2} - \underbrace{O(\sigma^2)}_{\to 0}.
    \end{equation}
    Thus, in the limit $\sigma \to 0$, the bias term vanishes, allowing us to replace the expected conditional variance with the global variance:
    \begin{equation}
        \mathbb{E}_x [\text{Var}[\eta_k | x]] = \text{Var}[\eta_k] \quad(\sigma \to 0).
    \end{equation}

    Substitude this into Eq~\eqref{eq:tmp}, we finally prove that
    \begin{equation}
    \begin{aligned}
    \label{eq:another_eic_eq}
    \text{EIC} &= \frac{1}{2} \lim_{\sigma \to 0} \max_{k \in \mathcal{T}[f]} \log_{10} \frac{\text{Var}[\eta_k]}{\sigma^2} \\
    &= \lim_{\sigma \to 0} \max_{k \in \mathcal{T}[f]} \log_{10} \sqrt{\frac{\text{Var}[\eta_k]}{\sigma^2}} \\
    &= \lim_{\sigma \to 0} \max_{k \in \mathcal{T}[f]} \log_{10} \frac{\sqrt{\text{Var}[\eta_k]}}{\sqrt{\text{Var}[\epsilon_k]}} \\
    &= \lim_{\sigma \to 0} \max_{k \in \mathcal{T}[f]} \log_{10} \frac{{\text{Std}[\eta_k]}}{{\text{Std}[\epsilon_k]}}.
    \end{aligned}
    \end{equation}
\end{proof}

\subsection{Invariance of EIC Under Unit Transformation}
\label{app:invariance_of_EIC}

\begin{proof}

Here we prove that the EIC is invariant under physical unit transformations (e.g., rescaling input variables from meters to centimeters). This invariance arises from the fact that the partial relative condition number $\kappa$ is intrinsically dimensionless, rendering the subsequent $s_k(x)$ and EIC metric independent of the chosen system of units. 

Specifically, let $[x]$ denote the physical dimension of a variable $x$ (e.g., Length, Time). For any computational operation $y_k = e(\{y_i\}_{i\in\mathcal{C}[k]})$, the partial derivative $\frac{\partial y_k}{\partial y_i}$ represents the rate of change of the output with respect to the input, and thus carries the dimension $[y_k][y_i]^{-1}$. The normalization term $\frac{y_i}{y_k}$ carries the dimension $[y_i][y_k]^{-1}$.

Multiplying these terms to obtain $\kappa_{k,i}$:
\begin{equation}
[\kappa_{k,i}] = \left[\frac{y_i}{y_k}\right] \cdot \left[\frac{\partial y_k}{\partial y_i}\right] = ([y_i][y_k]^{-1}) \cdot ([y_k][y_i]^{-1}) = 1.
\end{equation}
This confirms that $\kappa_{k,i}$ is a dimensionless quantity (a pure number).

Physical unit transformations (e.g., $x \to \lambda x$) act as linear scalings on the numerical values of dimensional quantities. However, dimensionless quantities are, by definition, invariant to such scaling because the scaling factors in the numerator and denominator cancel out exactly. Thus, the value of $\kappa_{k,i}$ remains constant regardless of the unit system.

As established in Proposition \ref{prop:s_k(x)}, the pointwise variance amplification $s_k(x)$ is computed recursively using solely the invariant $\kappa$ terms and the constant $1$. Consequently, both $s_k(x)$ and the final EIC derived from it are invariant under physical unit transformations.
\end{proof}

\subsection{Calculation of EIC}
\label{app:calculation_of_EIC}

Here, we use a simple example to illustrate how EIC is calculated and demonstrate its ability to assess the structural stability of formulas. Specifically, consider the identity $f(x) = x$ versus its mathematically equivalent but redundant form $g(x) = \ln(\exp(x))$. For the concise form $f(x)$, the pointwise amplification is constantly $s_f(x) = 1$. Thus, we have $\bar{s}_f = 1$ and $\text{EIC}(f) = \log_{10} 1 = 0$, suggesting no cumulated information loss during the calculation. 
For the redundant form $g(x)$, however, the intermediate node $u=\exp(x)$ has a condition number $\kappa_{u, x} = x$, leading to $s_u^2(x) = 1 + x^2$. The final node $y=\ln(u)$ has $\kappa_{y, u} = 1/x$, yielding $s_y^2(x) = 1 + (1/x)^2 (1+x^2) = 2 + x^{-2}$. 
When inputs $x$ follow a distribution with a large mean (e.g., $x \sim \mathcal{N}(\mu, 1)$ with $|\mu| \gg 1$), the intermediate node $u$ dominates the instability ($\bar{s}_u^2 = \mathbb{E}[s_u^2(x)] = 2 + \mu^2 \gg \bar{s}_y^2 = \mathbb{E}[s_{y}^2(x)] = 2 + \mathbb{E}[x^{-2}]$), yielding:
\begin{equation}
    \text{EIC}(g) = \log_{10} \bar{s}_u = \log_{10} \sqrt{2 + \mu^2} \approx \log_{10} \mu,
\end{equation}
which is larger than $\text{EIC}(f) = 0$ since $\mu \gg 1$. This result aligns with intuition: when inputs are distributed in a numerically large region, the exponentiation operation induces severe structural instability. Even though the subsequent logarithm restores the value to the correct range, the precision of the intermediate result remains hypersensitive to microscopic perturbations in $x$. This example highlights that EIC assesses a formula not merely as a functional mapping (Input $\to$ Output), but as an \textit{information processing system}. By evaluating the stability of the entire computational graph, EIC discerns the structural rationality of the representation, allowing it to identify the superior, robust realization, even among mathematically equivalent candidates.

This computation process can be formally described by pseudocode Algorithm~\ref{alg:calculate_eic}. 
It is worth noting that although Algorithm~\ref{alg:calculate_eic} requires a differential engine capable of calculating conditional partial derivatives, EIC can be estimated based on Eq~\eqref{eq:intuitive_of_EIC}. Specifically, this estimation strategy chooses a small enough $\sigma$ and traverses the symbolic tree in post-order, maintaining two parallel computation paths to simulate the noise propagation process: a ``clean'' reference path computed using standard double-precision arithmetic, and a ``noisy'' perturbed path where multiplicative noise $\epsilon \sim \mathcal{N}(0, \sigma^2)$ is injected at every operation node. Consequently, the instability factor $\bar{s}_k$ is estimated via the variance of the relative error between these two paths, which is then mapped to the EIC value according to Eq.~\eqref{eq:EIC_def}. While this stochastic approach shares the same time complexity $O(N)$ as the analytical method (where $N$ is the number of nodes), it is significantly easier to integrate into existing evaluation pipelines as it requires only standard forward passes. We detailedly describe this estimation strategy in Algorithm~\ref{alg:estimate_eic}.

\begin{algorithm}[tb]
    \caption{CalculateEIC (Recursively calculate partial relative condition number)}
    \label{alg:calculate_eic}
    \begin{algorithmic}
    \STATE {\bfseries Input:} Formula $f$ and data distribution $\mathcal{D}$ with $N$ samples.
    \STATE {\bfseries Output:} Variance amplification factor $s(x)$ and output value $y(x)$ at root node, as well as EIC value of $f$.
    \STATE
    \STATE $k \gets f.\text{root}$ \quad \% Start from the root node
    \IF{$k$ is a Leaf Node (Variable or Constant)}
        \STATE $y_k(x) \gets \text{Evaluate}(k, \mathcal{D})$ \quad \% Get values of $k$ in $\mathcal{D}$
        \STATE $s_k^2(x) \gets 1$
        \STATE $\bar{s}_k^2 \gets \mathbb{E}_{x\sim\mathcal{D}} [s_k^2(x)]$ \quad \% $\bar{s}_k=1$ for leaf nodes
        \STATE $\text{EIC}_k \gets \log_{10} \bar{s}_k$ \quad \% $\text{EIC}=0$ for leaf nodes
        \STATE {\bfseries return} $(s_k(x), y_k(x), \text{EIC}_k)$
    \ELSE
        \STATE $\text{EIC}_k \gets 0$
        \STATE $e_k \gets k.\text{operator}$
        \FOR{each child $i \in \mathcal{C}[k]$}
            \STATE $f' \gets \text{Subtree rooted at } i$
            \STATE $(s_i(x), y_i(x), \text{EIC}_i) \gets \text{CalculateEIC}(f', \mathcal{D})$
            \STATE $\text{EIC}_k \gets \max\{\text{EIC}_k, \text{EIC}_i\}$
        \ENDFOR
        \STATE $y_k(x) \gets e_k(\{y_{i}\}_{i\in\mathcal{C}[k]})$
        \FOR{each child $i \in \mathcal{C}[k]$}
            \STATE $\kappa_{k, i}(x) \gets \frac{y_i(x)}{y_k(x)} \frac{\partial e_{k}}{\partial y_i} |_{x\sim\mathcal{D}}$ %
        \ENDFOR
        \STATE $s_k^2(x) \gets 1 + \sum_{i \in \mathcal{C}[k]} \kappa_{k,i}^2 (x) s_i^2(x)$
        \STATE $\bar{s}_k^2 \gets \mathbb{E}_{x\sim\mathbb{E}} [s_k^2(x)]$
        \STATE $\text{EIC}_k \gets \max\{\text{EIC}_k, \log_{10} \bar{s}_k\}$
        \STATE {\bfseries return} $(s_k(x), y_k(x), \text{EIC}_k)$
    \ENDIF
    \end{algorithmic}
\end{algorithm}

\begin{algorithm}[tb]
    \caption{EstimateEIC (Recursively add stochastic perturbation)}
    \label{alg:estimate_eic}
    \begin{algorithmic}
    \STATE {\bfseries Input:} Formula $f$ and data distribution $\mathcal{D}$ with $N$ samples.
    \STATE {\bfseries Parameter:} A small enough noise scale $\sigma$.
    \STATE {\bfseries Output:} Noisy output $\tilde{y}(x)$ and clean output $y(x)$ at root node, as well as EIC value of $f$.
    \STATE
    \STATE $k \gets f.\text{root}$ \quad \% Start from the root node
    \IF{$k$ is a Leaf Node (Variable or Constant)}
        \STATE $y_k(x) \gets \text{Evaluate}(k, \mathcal{D})$ \quad \% Get values of $k$ in $\mathcal{D}$
        \STATE Generate $\epsilon_k \sim \mathcal{N}(0, \sigma^2)$ of size $N$
        \STATE $\tilde{y}_k(x) \gets y_k(x) \times (1 + \epsilon_k)$
        \STATE $\eta_k \gets \frac{\tilde{y}_k(x) - y_k(x)}{y_k(x)}$ \quad \% $\eta_k = \epsilon_k$ for leaf nodes
        \STATE $\text{EIC}_k \gets \log_{10} \sqrt{\text{Var}[\eta_k] / \sigma^2}$ \quad \% $\text{EIC}=0$ for leaf nodes
        \STATE {\bfseries return} $(\tilde{y}_k(x), y_k(x), \text{EIC}_k)$
    \ELSE
        \STATE $\text{EIC}_k \gets 0$
        \STATE $e_k \gets k.\text{operator}$
        \FOR{each child $i \in \mathcal{C}[k]$}
            \STATE $f' \gets \text{Subtree rooted at } i$
            \STATE $(\tilde{y}_i(x), y_i(x), \text{EIC}_i) \gets \text{EstimateEIC}(f', \mathcal{D}, \sigma)$
            \STATE $\text{EIC}_k \gets \max\{\text{EIC}_k, \text{EIC}_i\}$
        \ENDFOR
        \STATE $y_k(x) \gets e_k(\{y_{i}\}_{i\in\mathcal{C}[k]})$
        \STATE Generate $\epsilon_k \sim \mathcal{N}(0, \sigma^2)$ of size $N$
        \STATE $\tilde{y}_k(x) \gets e_k(\{\tilde{y}_{i}\}_{i\in\mathcal{C}[k]}) \times (1 + \epsilon_k)$
        \STATE $\eta_k \gets \frac{\tilde{y}_k(x) - y_k(x)}{y_k(x)}$
        \STATE $\text{EIC}_k \gets \max\{\text{EIC}_k, \log_{10} \sqrt{\text{Var}[\eta_k] / \sigma^2}\}$   \% Eq.~\eqref{eq:intuitive_of_EIC}
        \STATE {\bfseries return} $(\tilde{y}_k(x), y_k(x), \text{EIC}_k)$
    \ENDIF
    \end{algorithmic}
\end{algorithm}

\subsection{Interpretations for EIC}
\label{app:interprelation_for_EIC}

\noindent\textbf{Numerical Precision Perspective.}
EIC measures the stability of a formula's computational structure when multiplicative noise of a given intensity is introduced, which, from the perspective of numerical precision, corresponds to a specified number of significant digits.

According to quantization theory, the operation of rounding a value to retain $N$ significant decimal digits is statistically equivalent to injecting a multiplicative noise with variance $\sigma^2 \approx 10^{-2N}$, or $\sigma \approx 10^{-N}$~\citep{widrow1996statistical}.
Based on this equivalence, the injected noise $\epsilon_k$ with standard error $\text{Std}[\epsilon_k] = \sigma$ in Assumption~\ref{ass:noise_model} can be considered as the operation of rounding with a precision of $N \approx -\log_{10}(\text{Std}[\epsilon_k])$ significant digits.~\footnote{Although the rounding error typically follows a Uniform distribution, our derivation relies solely on the second moment. Therefore, the number of digits $N$ (and $M_k$) discussed here is defined in the \textit{variance-equivalent} sense, mapping the noise intensity to the precision level that would produce such variance regardless of the specific underlying distribution.} On the other hand, the accumulated relative (i.e., multiplicative) error $\eta_k$ implies that only $M_k \approx -\log_{10}(\text{Std}[\eta_k])$ effective significant digits remain at node $k$ due to the accumulation of injected noise.
Using this relationship, we can formally link EIC to the loss of precision. Specifically, in the high-precision limit ($\sigma \to 0$, corresponding to calculation precision $N \to \infty$), while the effective precision $M_k$ at any node $k$ increases with $N$ indefinitely, it asymptotically lags behind $N$ by a constant margin. We define the maximum inherent precision loss induced by the computational structure as:
\begin{equation}
    \Delta_{\max} \triangleq \lim_{N \to \infty} (N - \min_{k \in \mathcal{T}} M_k).
\end{equation}
which represents the maximum asymptotic gap between the applied computational precision and the effective precision achievable at any node $K$.
Based on the relationship between variance and precision, the EIC equals this loss:
\begin{equation}
    \text{EIC} = \Delta_{\max}.
\end{equation}
This indicates that the EIC measures the structural instability in terms of lost significant digits.

\begin{proof}
    To proof this, we can substitute the relationship $N \approx -\log_{10}(\text{Std}[\epsilon_k])$ and $M_k \approx -\log_{10}(\text{Std}[\eta_k])$ into Eq~\eqref{eq:intuitive_of_EIC}, obtaining:
    \begin{equation}
    \begin{aligned}
        \text{EIC} 
        &= \lim_{\sigma \to 0} \max_{k \in \mathcal{T}[f]} \log_{10} \frac{\text{Std}[\eta_k]}{\text{Std}[\epsilon_k]} \\
        &= \lim_{\sigma \to 0} \max_{k \in \mathcal{T}[f]} (\log_{10} \text{Std}[\eta_k] - \log_{10} \text{Std}[\epsilon_k]) \\
        &= \lim_{\sigma \to 0} \max_{k \in \mathcal{T}} ((-M_k) - (-N)) \\
        &= \lim_{\sigma \to 0} \max_{k \in \mathcal{T}} (N - M_k) \\
        &= \lim_{\sigma \to 0} (N - \min_{k \in \mathcal{T}} M_k) \\
        \label{eq:eic_limit_form}
    \end{aligned}
    \end{equation}
    This completes the proof.
\end{proof}

\noindent\textbf{Signal Processing Perspective.}
EIC can alternatively be interpreted through the perspective of signal processing. By treating the computational graph as a signal transmission channel, we can analyze how the quality of the ``information signal'' degrades as it propagates from input to output.

For any node $k$, we define the ``signal'' as the noise-free output value $y_k$, and the ``noise'' as the deviation caused by the accumulated relative error $\tilde{y}_k - y_k = y_k \eta_k$. The local Signal-to-Noise Ratio (SNR) is defined as the ratio of signal power to noise power:
\begin{equation}
    \text{SNR}_{k} \triangleq \frac{\text{Signal Power}}{\text{Noise Power}} = \frac{y_k^2}{\text{Var}[y_k \eta_k]} = \frac{y_k^2}{y_k^2 \text{Var}[\eta_k]} = \frac{1}{\text{Var}[\eta_k]}.
\end{equation}
Similarly, the intrinsic SNR is.
\begin{equation}
    \text{SNR}_{\text{intrinsic}} \triangleq \frac{y_k^2}{\text{Var}[y_k \epsilon_k]} = \frac{1}{\sigma^2}
\end{equation}
To quantify the degradation of signal quality, we utilize the \textit{Noise Figure} $F_k \triangleq 10 \log_{10} \frac{\text{SNR}_{\text{intrinsic}}}{\text{SNR}_{k}}$, which measures the ratio of input SNR to output SNR.

From this perspective, we can interpret EIC as the noise figure (scaled by $\frac{1}{20}$ at the bottleneck nodes:
\begin{equation}
\text{EIC} = \frac{1}{20} \lim_{\sigma \to 0} \max_{k\in\mathcal{T}[f]} F_k
\end{equation}
\begin{proof}
Begin from the Eq~\eqref{eq:intuitive_of_EIC}, we have
\begin{equation}
\begin{aligned}
    \text{EIC} 
    &= \lim_{\sigma \to 0} \max_{k \in \mathcal{T}[f]} \log_{10} \frac{\text{Std}[\eta_k]}{\text{Std}[\epsilon_k]} \\
    &= \frac{1}{2}\lim_{\sigma \to 0} \max_{k \in \mathcal{T}[f]} \log_{10} \frac{\text{Var}[\eta_k]}{\sigma^2} \\
    &= \frac{1}{2}\lim_{\sigma \to 0} \max_{k \in \mathcal{T}[f]} \log_{10} \frac{\text{SNR}_{\text{intrinsic}}}{\text{SNR}_{k}} \\
    &= \frac{1}{20}\lim_{\sigma \to 0} \max_{k \in \mathcal{T}[f]} F_k.
\end{aligned}
\end{equation}
This completes the proof.
\end{proof}

\section{Methodology Details}

\subsection{Enhance Search-based Methods with EIC}
\label{app:eic_for_search}

To demonstrate that EIC, as a criterion for evaluating unreasonable structures in formulas, can serve as an effective guidance signal for improving symbolic regression, we integrated it into various classical heuristic search algorithms as an auxiliary search objective. Specifically, we focused on two representative approaches: genetic programming (GP) and Monte Carlo tree search (MCTS).
\begin{itemize}
\item Genetic programming maintains a population of candidate formulas, where each individual represents a candidate formula. New individuals are generated by crossover or mutation of high-fitness candidates, while low-fitness candidates are eliminated to increase the overall quality of the population. A common fitness function is defined as
\begin{equation}
\label{eq:fitness}
f(C, \text{NMSE}; \eta) = {\eta^{C}}/{(1+\text{NMSE})},
\end{equation}
where \textit{Complexity} measures the structural size of the formula, NMSE denotes the normalized mean squared error ($\text{MSE}/\text{Var}(y)$), and $\eta < 1$ is a regularization constant. This formulation penalizes formulas with large errors or excessive complexity, thereby guiding the search toward accurate and compact formulas. It is worth noting that, when calculating the MSE, we first decompose formulas into additive terms and then apply linear regression to fit the coefficients for these terms. This approach integrates the idea of SINDy and effectively improves the performance of algorithms. 
\item MCTS constructs a search tree where each node represents a candidate formula. Child nodes are those formulas that can be obtained by mutating parent formulas, and their reward values are computed according to Equation~\eqref{eq:fitness}. In each search iteration, the algorithm starts from the root node and selects a promising leaf node based on its upper confidence bound (UCB) score, which balances average reward and visitation counts. The chosen leaf is then expanded through mutation, and the resulting reward is backpropagated to update the average reward and visitation counts of its ancestors. This process iteratively guides the search toward structurally and numerically promising formulas.
\end{itemize}
The proposed EIC can be easily incorporated into both algorithms by augmenting their fitness or reward functions. Specifically, we adopt
$$
\text{Fitness}_{\alpha} = \eta^{\text{Complexity}} / (1+\text{NMSE}) - \alpha \cdot \text{EIC}
$$
as the modified fitness and reward functions, where $\alpha > 0$ penalizes formulas with higher EIC, thereby steering the search away from structurally unreasonable solutions.
In this work, we use $\eta = 0.999$ in \eqref{eq:fitness} as suggested by \citet{yu2025symbolic} and \citet{sun2022symbolic}. For the choice of $\alpha$, we note that the first term of \eqref{eq:fitness} has a value range of 0 to 1, while, as shown in Figure~\ref{fig:EIC_compare}, EIC ranges from 0 to 10. Since EIC serves as an auxiliary search objective, its weight should generally be an order of magnitude smaller than that of the primary objective. Accordingly, we set $\alpha=0.01$ for MCTS. For the genetic algorithm, we observed that it is more sensitive to the auxiliary objective. Therefore, we chose a smaller value of $\alpha=0.002$ to optimally balance the algorithm’s ability to discover formulas with low EIC while still ensuring high $R^2$ accuracy.

\subsection{Enhance Generative Methods with EIC}
\label{app:eic_for_generative}

To demonstrate that EIC can enhance the sample efficiency and performance of pretraining-based symbolic regression methods, we focus on three representative approaches: E2ESR\citep{kamienny2022end}, SNIP\citep{meidani2023snip}, and SR4MDL\citep{yu2025symbolic}, spanning the progression of this research line from earlier efforts to more recent advances in the last year. E2ESR is trained to predict formula tokens directly from $(X, y)$ data pairs; SNIP extends E2ESR with a CLIP-inspired contrastive loss; and SR4MDL further modifies SNIP by changing the prediction target to the minimum description length of formulas to guide the search. All three approaches rely on the formula generation algorithm proposed by \citet{lample2019deep}, which generates formulas with random and diverse forms to pretrain generative models.

For each method, we pretrained the model on randomly generated formulas until it achieved the performance reported in the corresponding original papers, and then repeated the pretraining process using a filtered dataset where formulas with $\text{EIC} > 2.0$ are discarded, ensuring that only formulas with reasonable structures were included in pre-training. For E2ESR and SNIP, since they directly predict formula tokens from data, we evaluated their performance by the $R^2$ scores of the formulas they generated on the Feynman dataset. For SR4MDL, which instead predicts the minimum description length of formulas, we evaluated its performance using the mean absolute error (MAE) and root mean squared error (RMSE) of predicted formula lengths on the Feynman dataset. This setup allowed us to compare the number of training samples required to reach the same target performance under both settings. All experiments used their official implementations, with hyperparameters selected via a grid search over batch sizes $\in\{32, 64, 128, 256\}$ and learning rates $\in\{10^{-5}, 10^{-4}, 10^{-3}\}$. The hyperparameters were selected under the unfiltered training condition and then kept fixed when training with the filtered dataset.

We also considered a recently proposed data construction strategy baseline, PhyE2E\citep{ying2025neural}, which leverages LLMs fine-tuned on physical equations to generate ``look-physical'' formulas with unit constraints for pretraining generative models. We used 180k formulas provided by PhyE2E to train the model. To avoid unfair comparisons due to limited sample size, we used a mixed sampling strategy, which is to generate random formulas with a probability of 0.9 and sample from PhyE2E formulas with a probability of 0.1.
We used the 180k formulas provided by PhyE2E to pretrain the model and compared the results against those obtained with the EIC-filtered dataset. To ensure a fair comparison with our approach and to mitigate the limited size of the PhyE2E corpus, we adopted a hybrid sampling scheme in which, at each training step, a formula was drawn from the PhyE2E dataset with probability $0.1$ and from the model’s own random generator with probability $0.9$.

\section{Detailed Experimental Results}

\subsection{Structural Instability of LLM-Generated Formulas}
\label{app:llm_sr}

In addition to search-based and generative symbolic regression methods, we further investigate the structural properties of formulas generated by large language models (LLMs) in LLM-SRBench \citep{shojaee2025llm}. 
This benchmark includes two categories of LLM-generated formulas: \emph{Transform} and \emph{Synth}. 
The Transform subset is obtained by applying symbolic transformations (e.g., inversion) to real physical formulas, while the Synth subset is constructed via \emph{LLM-assisted synthesis of known and novel terms}, where LLMs combine established scientific components with newly synthesized symbolic terms.

\begin{figure}[htbp]
    \centering
    \includegraphics[width=0.4\linewidth]{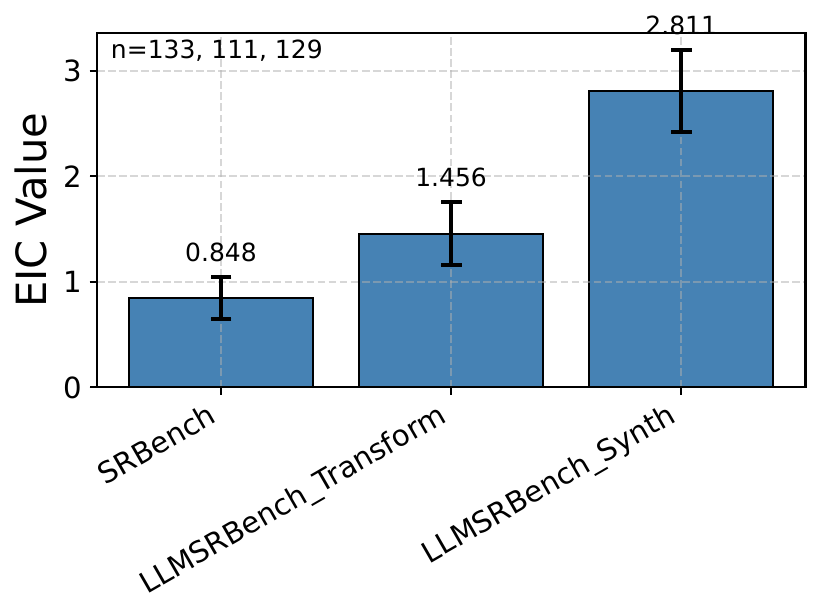}
    \caption{
        \textbf{Average EIC comparison between real physical formulas and LLM-generated formulas.}
        Error bars denote 95\% confidence intervals.
    }
    \label{fig:llm_eic_bar}
\end{figure}

\begin{figure}[htbp]
    \centering
    \includegraphics[width=0.4\linewidth]{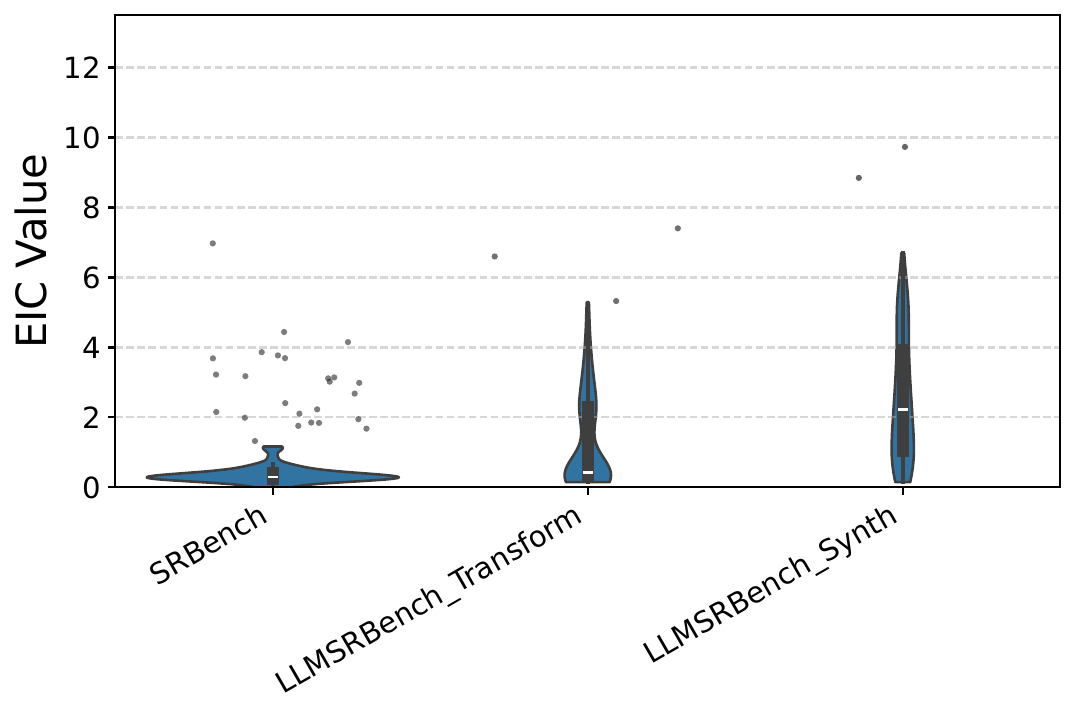}
    \caption{
        \textbf{Violin plots of EIC distributions for SRBench physical formulas, Transform, and Synth subsets.}
        Scattered points indicate outliers.
    }
    \label{fig:llm_eic_violin}
\end{figure}

Figure~\ref{fig:llm_eic_bar} and Figure~\ref{fig:llm_eic_violin} present the EIC distributions of formulas from SRBench physical formulas, Transform, and Synth subsets, respectively. 
Both LLM-generated subsets exhibit substantially higher EIC values than real physical formulas from SRBench, indicating a higher degree of structural instability. 
In particular, formulas in the Synth subset show the highest EIC distribution, suggesting that LLM-synthesized symbolic expressions tend to involve structurally fragile or numerically unstable constructions, despite their surface-level plausibility.

These results lead to three key observations.
First, formulas synthesized by LLMs (Synth) exhibit significantly higher EIC values than real physical formulas, indicating that both LLM-generated symbolic expressions and LLM-based symbolic regression methods inherit the challenge of structural instability.
Second, formulas derived from physical transformations (Transform) show notably lower EIC than Synth formulas, providing further evidence that symbolic expressions rooted in natural scientific laws differ fundamentally from those synthesized purely by LLMs.
Third, LSR-Transform formulas still exhibit higher EIC than the original SRBench physical formulas.
We attribute this increase to the fact that some physical equations are not naturally invertible from a causal or physical perspective. When such equations are algebraically inverted to construct LSR-Transform tasks, the resulting expressions may involve divisions by quantities that can become arbitrarily small over the data domain, thereby introducing numerically and structurally unstable forms.

\subsection{Enhance Search-based Methods with EIC}
\label{app:search_results}

\begin{figure*}[htbp]
    \centering
    \includegraphics[width=\linewidth]{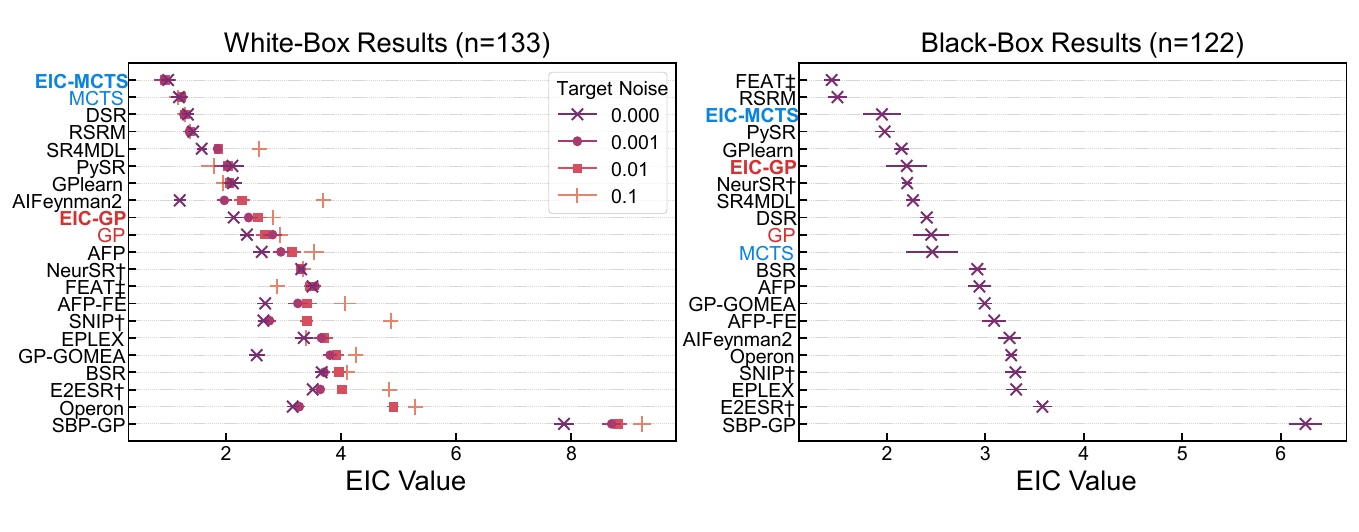}
    \caption{
        \textbf{EIC distributions of formulas discovered by baseline methods and our EIC-enhanced variants.} 
        The left panel presents results on 133 white-box problems under varying noise levels ($\{0, 10^{-3}, 10^{-2}, 10^{-1}\}$), while the right panel presents results on 122 black-box problems. The markers represent the mean EIC values across all problems, and error bars indicate 95\% confidence intervals. Notably, the EIC-enhanced methods (highlighted in bold color) consistently exhibit lower EIC values compared to their vanilla counterparts, demonstrating that incorporating EIC effectively steers the search toward structurally rational formulas.
    }
    \label{fig:app_eic_dist}
\end{figure*}

Figure~\ref{fig:app_eic_dist} illustrates the distribution of Effective Information Criterion (EIC) values for formulas discovered by all benchmarked methods. The left panel depicts the distributions on white-box problems across varying noise levels. Notably, the integration of EIC consistently drives both MCTS and GP toward lower EIC values, aligning them more closely with the distribution of ground-truth physical formulas compared to their vanilla versions. This confirms that EIC effectively suppresses structurally unreasonable and numerically unstable patterns. The right panel presents the results on black-box problems. Here, incorporating EIC leads to a substantial reduction in the prevalence of high-EIC structures for both MCTS and GP. Crucially, this reduction is more pronounced compared to the white-box setting, suggesting that in the absence of a known underlying physical law, standard SR methods are prone to overfitting via overly complex and unstable functional forms. EIC effectively mitigates this tendency, enforcing structural plausibility even when the ground truth is unknown.

Corresponding to Figures~\ref{fig:merged_results} (in the main text) as well as Figure~\ref{fig:app_eic_dist}, we provide the raw experimental results for 17 baseline methods and our four methods under four noise levels on white-box data and on black-box data, as shown in Tables \ref{tab:whitebox0}, \ref{tab:whitebox0001}, \ref{tab:whitebox001}, \ref{tab:whitebox01}, and \ref{tab:blackbox}, respectively. The tables report formula accuracy, complexity, running duration, and the average EIC of the resulting formulas.

\begin{table*}[htbp]
  \centering
  \caption{Whitebox results at noise-free condition}
    \begin{tabular}{cccccc}
    \toprule
    \textbf{Type} & \textbf{algorithm} & \textbf{$R^2>0.99$ $\uparrow$} & \textbf{complexity$\downarrow$} & \textbf{duration$\downarrow$} & \textbf{EIC$\downarrow$} \\
    \midrule
    Regression & \textbf{FEAT} & 0.621\tiny{(±0.031)} & 195.3\tiny{(±8.5)} & 2269\tiny{(±1.9e+02)} & 3.504\tiny{(±0.22)} \\
    \midrule
    \multirow{3}[2]{*}{Generative} & \textbf{E2ESR} & 0.2773\tiny{(±0.03)} & 89.63\tiny{(±2.1)} & 4.024\tiny{(±0.14)} & 3.509\tiny{(±0.066)} \\
          & \textbf{NeurSR} & 0.07681\tiny{(±0.014)} & 31.42\tiny{(±0.26)} & 23.19\tiny{(±0.39)} & 3.307\tiny{(±0.049)} \\
          & \textbf{SNIP} & 0.1541\tiny{(±0.019)} & 25.2\tiny{(±0.39)} & 1.845\tiny{(±0.07)} & 2.652\tiny{(±0.1)} \\
    \midrule
    \multirow{19}[3]{*}{Search} & \textbf{AFP} & 0.551\tiny{(±0.03)} & 37.01\tiny{(±1.2)} & 3282\tiny{(±2.4e+02)} & 2.622\tiny{(±0.15)} \\
          & \textbf{AFP-FE} & 0.7154\tiny{(±0.025)} & 40.63\tiny{(±1.3)} & 17090\tiny{(±750)} & 2.684\tiny{(±0.14)} \\
          & \textbf{AIFeynman2} & 0.8487\tiny{(±0.02)} & 113.2\tiny{(±21)} & 844.2\tiny{(±1.9e+02)} & 1.2\tiny{(±0.2)} \\
          & \textbf{BSR} & 0.249\tiny{(±0.026)} & 26.95\tiny{(±0.62)} & 29310\tiny{(±790)} & 3.663\tiny{(±0.11)} \\
          & \textbf{DSR} & 0.3365\tiny{(±0.029)} & 14.94\tiny{(±0.49)} & 1746\tiny{(±1.8e+02)} & 1.343\tiny{(±0.1)} \\
          & \textbf{EPLEX} & 0.7375\tiny{(±0.027)} & 52.64\tiny{(±1.1)} & 11450\tiny{(±690)} & 3.355\tiny{(±0.15)} \\
          & \textbf{GP-GOMEA} & 0.8808\tiny{(±0.02)} & 34.77\tiny{(±0.97)} & 4786\tiny{(±4.4e+02)} & 2.537\tiny{(±0.14)} \\
          & \textbf{GPlearn} & 0.468\tiny{(±0.03)} & 67.71\tiny{(±16)} & 3606\tiny{(±3.6e+02)} & 2.126\tiny{(±0.15)} \\
          & \textbf{Operon} & 0.9392\tiny{(±0.013)} & 68.73\tiny{(±1.4)} & 1947\tiny{(±64)} & 3.162\tiny{(±0.11)} \\
          & \textbf{PySR} & 0.6762\tiny{(±0.064)} & 9.219\tiny{(±0.46)} & 791.6\tiny{(±44)} & 2.114\tiny{(±0.2)} \\
          & \textbf{RSRM} & 0.2455\tiny{(±0.025)} & 13.4\tiny{(±0.37)} & 116.9\tiny{(±3)} & 1.435\tiny{(±0.092)} \\
          & \textbf{SBP-GP} & 0.9365\tiny{(±0.015)} & 513.4\tiny{(±12)} & 2.768e+04\tiny{(±2.4e+02)} & 7.872\tiny{(±0.34)} \\
          & \textbf{SR4MDL} & 0.6271\tiny{(±0.026)} & 21.7\tiny{(±0.79)} & 470\tiny{(±34)} & 1.58\tiny{(±0.086)} \\
\cmidrule{2-6}          & \textbf{GP} & 0.7215\tiny{(±0.029)} & 34.61\tiny{(±1.7)} & 1728\tiny{(±67)} & 2.367\tiny{(±0.12)} \\
          & \textbf{EIC-GP} & 0.7288\tiny{(±0.038)} & 32.47\tiny{(±2.2)} & 1718\tiny{(±90)} & 2.136\tiny{(±0.15)} \\
          & $\Delta(\%)$      & 1\%   & -7\%  & -0.50\% & -9.80\% \\
\cmidrule{2-6}          & \textbf{MCTS} & 0.6917\tiny{(±0.046)} & 17.68\tiny{(±1.2)} & 10580\tiny{(±1000)} & 1.185\tiny{(±0.14)} \\
          & \textbf{EIC-MCTS} & 0.7107\tiny{(±0.045)} & 17.64\tiny{(±1.2)} & 12080\tiny{(±1.1e+03)} & 0.9983\tiny{(±0.12)} \\
          & $\Delta(\%)$      & 2.70\% & -0.20\% & 14\%  & -16\% \\
              \toprule

    \end{tabular}%
  \label{tab:whitebox0}%
\end{table*}%

\begin{table*}[htbp]
  \centering
  \caption{Whitebox Results at 0.001 noise}
    \begin{tabular}{cccccc}
    \toprule
    \textbf{Type} & \textbf{algorithm} & \textbf{$R^2>0.99$ $\uparrow$} & \textbf{complexity$\downarrow$} & \textbf{duration$\downarrow$} & \textbf{EIC$\downarrow$} \\
    \midrule
    Regression & \textbf{FEAT} & 0.6236	\tiny{(±0.03)} & 186.4	\tiny{(±7.2)} & 1598	\tiny{(±89)} & 3.556	\tiny{(±0.18)} \\
    \midrule
    \multirow{3}[2]{*}{Generative} & \textbf{E2ESR} & 0.2595	\tiny{(±0.03)} & 89.56	\tiny{(±2.1)} & 3.846	\tiny{(±0.13)} & 3.641	\tiny{(±0.07)} \\
          & \textbf{NeurSR} & 0.0767	\tiny{(±0.01)} & 31.34	\tiny{(±0.26)} & 23.52	\tiny{(±0.39)} & 3.309	\tiny{(±0.05)} \\
          & \textbf{SNIP} & 0.1579	\tiny{(±0.02)} & 25.2	\tiny{(±0.39)} & 1.865	\tiny{(±0.07)} & 2.759	\tiny{(±0.11)} \\
    \midrule
    \multirow{19}[4]{*}{Search} & \textbf{AFP} & 0.5646	\tiny{(±0.03)} & 39.27	\tiny{(±1)} & 3327	\tiny{(±104)} & 2.954	\tiny{(±0.14)} \\
          & \textbf{AFP-FE} & 0.7308	\tiny{(±0.02)} & 46.71	\tiny{(±1.1)} & 24216	\tiny{(±511)} & 3.252	\tiny{(±0.17)} \\
          & \textbf{AIFeynman2} & 0.8305	\tiny{(±0.02)} & 118.7	\tiny{(±20)} & 575.1	\tiny{(±27)} & 1.974	\tiny{(±0.25)} \\
          & \textbf{BSR} & 0.2515	\tiny{(±0.02)} & 27.12	\tiny{(±0.56)} & 29613	\tiny{(±1691)} & 3.709	\tiny{(±0.1)} \\
          & \textbf{DSR} & 0.3815	\tiny{(±0.03)} & 16.32	\tiny{(±0.47)} & 794.6	\tiny{(±27)} & 1.273	\tiny{(±0.09)} \\
          & \textbf{EPLEX} & 0.77	\tiny{(±0.02)} & 55.3	\tiny{(±0.8)} & 11057	\tiny{(±318)} & 3.66	\tiny{(±0.14)} \\
          & \textbf{GP-GOMEA} & 0.9038	\tiny{(±0.02)} & 44.79	\tiny{(±0.77)} & 2678	\tiny{(±197)} & 3.811	\tiny{(±0.13)} \\
          & \textbf{GPlearn} & 0.4809	\tiny{(±0.03)} & 57.12	\tiny{(±8.4)} & 3054	\tiny{(±234)} & 2.093	\tiny{(±0.13)} \\
          & \textbf{Operon} & 0.9538	\tiny{(±0.01)} & 69.39	\tiny{(±1.4)} & 1967	\tiny{(±70)} & 3.276	\tiny{(±0.10)} \\
          & \textbf{PySR} & 0.7914	\tiny{(±0.07)} & 9.381	\tiny{(±0.5)} & 658.6	\tiny{(±34)} & 2.029	\tiny{(±0.23)} \\
          & \textbf{RSRM} & 0.263	\tiny{(±0.03)} & 13.22	\tiny{(±0.47)} & 128.3	\tiny{(±2.5)} & 1.367	\tiny{(±0.12)} \\
          & \textbf{SBP-GP} & 0.9362	\tiny{(±0.01)} & 600.9	\tiny{(±9.4)} & 27934	\tiny{(±172)} & 8.697	\tiny{(±0.32)} \\
          & \textbf{SR4MDL} & 0.6099	\tiny{(±0.03)} & 26.8	\tiny{(±0.65)} & 540.5	\tiny{(±28)} & 1.855	\tiny{(±0.08)} \\
\cmidrule{2-6}          & \textbf{GP} & 0.691	\tiny{(±0.05)} & 35.65	\tiny{(±2.3)} & 1813	\tiny{(±103)} & 2.811	\tiny{(±0.16)} \\
          & \textbf{EIC-GP} & 0.7103	\tiny{(±0.04)} & 31.7	\tiny{(±2.2)} & 1703	\tiny{(±103)} & 2.394	\tiny{(±0.16)} \\
          &  $\Delta(\%)$     & +3\%   & -11\%  & -6.08\% & -14.82\% \\
\cmidrule{2-6}          & \textbf{MCTS} & 0.6404	\tiny{(±0.03)} & 19.3	\tiny{(±0.6)} & 3641	\tiny{(±469)} & 1.246	\tiny{(±0.09)} \\
          & \textbf{EIC-MCTS} & 0.6692	\tiny{(±0.08)} & 20.46	\tiny{(±1.8)} & 15228	\tiny{(±2045)} & 0.9275	\tiny{(±0.18)} \\
          & $\Delta(\%)$      & +4.49\% & +6.03\% & +318\% & -26\% \\
    \bottomrule
    \end{tabular}%
  \label{tab:whitebox0001}%
\end{table*}%

\begin{table*}[htbp]
  \centering
  \caption{Whitebox Results at 0.01 noise}
    \begin{tabular}{cccccc}
    \toprule
    \textbf{Type} & \textbf{algorithm} & \textbf{$R^2>0.99$ $\uparrow$} & \textbf{complexity$\downarrow$} & \textbf{duration$\downarrow$} & \textbf{EIC$\downarrow$} \\
    \midrule
    Regression & \textbf{FEAT} & 0.6198	\tiny{(±0.03)} & 159.3	\tiny{(±6.2)} & 1357	\tiny{(±73)} & 3.464	\tiny{(±0.2)} \\
    \midrule
    \multirow{3}[2]{*}{Generative} & \textbf{E2ESR} & 0.2388	\tiny{(±0.03)} & 97.61	\tiny{(±2.3)} & 4.175	\tiny{(±0.1)} & 4.015	\tiny{(±0.08)} \\
          & \textbf{NeurSR} & 0.0759	\tiny{(±0.01)} & 31.44	\tiny{(±0.25)} & 23.66	\tiny{(±0.38)} & 3.292	\tiny{(±0.04)} \\
          & \textbf{SNIP} & 0.1414	\tiny{(±0.02)} & 27.05	\tiny{(±0.37)} & 2.109	\tiny{(±0.08)} & 3.407	\tiny{(±0.11)} \\
    \midrule
    \multirow{19}[4]{*}{Search} & \textbf{AFP} & 0.5808	\tiny{(±0.03)} & 40.62	\tiny{(±1)} & 3666	\tiny{(±118)} & 3.152	\tiny{(±0.16)} \\
          & \textbf{AFP-FE} & 0.7262	\tiny{(±0.02)} & 47.12	\tiny{(±1.1)} & 25731	\tiny{(±414)} & 3.413	\tiny{(±0.17)} \\
          & \textbf{AIFeynman2} & 0.797	\tiny{(±0.02)} & 140.2	\tiny{(±23)} & 562.8	\tiny{(±28)} & 2.276	\tiny{(±0.27)} \\
          & \textbf{BSR} & 0.2685	\tiny{(±0.02)} & 29.19	\tiny{(±0.71)} & 29680	\tiny{(±1554)} & 3.961	\tiny{(±0.12)} \\
          & \textbf{DSR} & 0.3838	\tiny{(±0.03)} & 16.45	\tiny{(±0.48)} & 882.8	\tiny{(±33)} & 1.289	\tiny{(±0.09)} \\
          & \textbf{EPLEX} & 0.7792	\tiny{(±0.02)} & 53.9	\tiny{(±0.82)} & 9901	\tiny{(±240)} & 3.711	\tiny{(±0.15)} \\
          & \textbf{GP-GOMEA} & 0.9085	\tiny{(±0.02)} & 44.46	\tiny{(±0.72)} & 2777	\tiny{(±197)} & 3.919	\tiny{(±0.13)} \\
          & \textbf{GPlearn} & 0.4813	\tiny{(±0.03)} & 56.85	\tiny{(±11)} & 3084	\tiny{(±235)} & 2.064	\tiny{(±0.13)} \\
          & \textbf{Operon} & 0.9438	\tiny{(±0.01)} & 87.29	\tiny{(±0.42)} & 2835	\tiny{(±80)} & 4.908	\tiny{(±0.1)} \\
          & \textbf{PySR} & 0.7857	\tiny{(±0.07)} & 9.545	\tiny{(±0.5)} & 803.6	\tiny{(±22)} & 2.034	\tiny{(±0.22)} \\
          & \textbf{RSRM} & 0.2703	\tiny{(±0.03)} & 13.38	\tiny{(±0.37)} & 132.7	\tiny{(±2.3)} & 1.384	\tiny{(±0.09)} \\
          & \textbf{SBP-GP} & 0.9338	\tiny{(±0.01)} & 623	\tiny{(±9.2)} & 28074	\tiny{(±154)} & 8.804	\tiny{(±0.31)} \\
          & \textbf{SR4MDL} & 0.6084	\tiny{(±0.03)} & 26.6	\tiny{(±0.64)} & 805.6	\tiny{(±41)} & 1.876	\tiny{(±0.08)} \\
\cmidrule{2-6}          & \textbf{GP} & 0.6642	\tiny{(±0.05)} & 34.83	\tiny{(±2)} & 1784	\tiny{(±100)} & 2.684	\tiny{(±0.16)} \\
          & \textbf{EIC-GP} & 0.6884	\tiny{(±0.05)} & 33.65	\tiny{(±2.2)} & 1582	\tiny{(±104)} & 2.554	\tiny{(±0.16)} \\
          &  $\Delta(\%)$     & +4\%   & -3\%   & -11.31\% & -4.84\% \\
\cmidrule{2-6}          & \textbf{MCTS} & 0.6186	\tiny{(±0.03)} & 19.33	\tiny{(±0.63)} & 4158	\tiny{(±464)} & 1.226	\tiny{(±0.09)} \\
          & \textbf{EIC-MCTS} & 0.7083	\tiny{(±0.08)} & 18.99	\tiny{(±1.8)} & 15996	\tiny{(±1928)} & 0.9407	\tiny{(±0.2)} \\
          &  $\Delta(\%)$     & +14.51\% & -1.77\% & +285\% & -23\% \\
    \bottomrule
    \end{tabular}%
  \label{tab:whitebox001}%
\end{table*}%

\begin{table*}[htbp]
  \centering
  \caption{Whitebox Results at 0.1 noise}
    \begin{tabular}{cccccc}
    \toprule
    \textbf{Type} & \textbf{algorithm} & \textbf{$R^2>0.99$ $\uparrow$} & \textbf{complexity$\downarrow$} & \textbf{duration$\downarrow$} & \textbf{EIC$\downarrow$} \\
    \midrule
    Regression & \textbf{FEAT} & 0.4606	\tiny{(±0.03)} & 97.84	\tiny{(±3.8)} & 742.3	\tiny{(±32)} & 2.891	\tiny{(±0.17)} \\
    \midrule
    \multirow{3}[2]{*}{Generative} & \textbf{E2ESR} & 0.0176	\tiny{(±0.01)} & 103.3	\tiny{(±2.1)} & 6.119	\tiny{(±0.34)} & 4.841	\tiny{(±0.1)} \\
          & \textbf{NeurSR} & 0.0421	\tiny{(±0.01)} & 31.83	\tiny{(±0.24)} & 24.93	\tiny{(±0.39)} & 3.344	\tiny{(±0.04)} \\
          & \textbf{SNIP} & 0.0346	\tiny{(±0.01)} & 30.95	\tiny{(±0.35)} & 2.655	\tiny{(±0.1)} & 4.862	\tiny{(±0.11)} \\
    \midrule
    \multirow{19}[4]{*}{Search} & \textbf{AFP} & 0.5485	\tiny{(±0.03)} & 41.18	\tiny{(±0.99)} & 3485	\tiny{(±95)} & 3.535	\tiny{(±0.17)} \\
          & \textbf{AFP-FE} & 0.7262	\tiny{(±0.02)} & 49.1	\tiny{(±1)} & 26795	\tiny{(±319)} & 4.077	\tiny{(±0.19)} \\
          & \textbf{AIFeynman2} & 0.1949	\tiny{(±0.02)} & 157.3	\tiny{(±21)} & 629.5	\tiny{(±91)} & 3.695	\tiny{(±0.26)} \\
          & \textbf{BSR} & 0.22	\tiny{(±0.02)} & 31.01	\tiny{(±0.84)} & 31506	\tiny{(±2521)} & 4.104	\tiny{(±0.11)} \\
          & \textbf{DSR} & 0.3823	\tiny{(±0.03)} & 16.3	\tiny{(±0.47)} & 781	\tiny{(±24)} & 1.288	\tiny{(±0.09)} \\
          & \textbf{EPLEX} & 0.7585	\tiny{(±0.02)} & 46.46	\tiny{(±0.95)} & 9219	\tiny{(±192)} & 3.394	\tiny{(±0.15)} \\
          & \textbf{GP-GOMEA} & 0.8885	\tiny{(±0.02)} & 46.13	\tiny{(±0.7)} & 2886	\tiny{(±207)} & 4.258	\tiny{(±0.12)} \\
          & \textbf{GPlearn} & 0.4786	\tiny{(±0.03)} & 46.32	\tiny{(±6.3)} & 2715	\tiny{(±209)} & 1.952	\tiny{(±0.13)} \\
          & \textbf{Operon} & 0.8892	\tiny{(±0.02)} & 88.61	\tiny{(±0.32)} & 2768	\tiny{(±71)} & 5.29	\tiny{(±0.1)} \\
          & \textbf{PySR} & 0.6234	\tiny{(±0.08)} & 10.49	\tiny{(±0.78)} & 798	\tiny{(±17)} & 1.797	\tiny{(±0.23)} \\
          & \textbf{RSRM} & 0.2543	\tiny{(±0.03)} & 13.11	\tiny{(±0.34)} & 134.2	\tiny{(±2.1)} & 1.379	\tiny{(±0.09)} \\
          & \textbf{SBP-GP} & 0.8662	\tiny{(±0.02)} & 652	\tiny{(±8.9)} & 28173	\tiny{(±140)} & 9.227	\tiny{(±0.31)} \\
          & \textbf{SR4MDL} & 0.4717	\tiny{(±0.03)} & 29.81	\tiny{(±0.59)} & 833.9	\tiny{(±44)} & 2.577	\tiny{(±0.07)} \\
\cmidrule{2-6}          & \textbf{GP} & 0.3258	\tiny{(±0.05)} & 29.67	\tiny{(±1.7)} & 1157	\tiny{(±96)} & 2.938	\tiny{(±0.15)} \\
          & \textbf{EIC-GP} & 0.3182	\tiny{(±0.05)} & 29.97	\tiny{(±1.9)} & 861.7	\tiny{(±91)} & 2.823	\tiny{(±0.14)} \\
          &   $\Delta(\%)$    & -2\%  & +1\%  & -25.51\% & -3.92\% \\
\cmidrule{2-6}          & \textbf{MCTS} & 0.5064	\tiny{(±0.06)} & 18.77	\tiny{(±1.1)} & 2778	\tiny{(±325)} & 1.174	\tiny{(±0.17)} \\
          & \textbf{EIC-MCTS} & 0.5012 \tiny{(±0.06)} & 18.67 \tiny{(±1.3)} & 3005 \tiny{(±403)} & 1.092 \tiny{(±0.16)} \\
          & $\Delta(\%)$ & -1\%      & +0.5\%      & +8.17\%      & -6.98\%    \\
    \bottomrule
    \end{tabular}%
  \label{tab:whitebox01}%
\end{table*}%

\begin{table*}[htbp]
  \centering
  \caption{blackbox results}
    \begin{tabular}{ccccc}
    \toprule
    \textbf{Type} & \textbf{algorithm} & \textbf{$R^2$ $\uparrow$} & \textbf{complexity $\downarrow$} & \textbf{EIC $\downarrow$} \\
    \midrule
    Regression & \textbf{FEAT} & 0.7621	\tiny{(±0.01)} & 82.49	\tiny{(±3.3)}      & 1.441	\tiny{(±0.09)} \\
    \midrule
    \multirow{3}[2]{*}{Generative} & \textbf{E2ESR} & 0.3612	\tiny{(±0.02)} & 61.09	\tiny{(±1)}      & 3.581	\tiny{(±0.13)} \\
          & \textbf{NeurSR} & 0.1228	\tiny{(±0.01)} & 13.33	\tiny{(±0.12)}      & 2.208	\tiny{(±0.07)} \\
          & \textbf{SNIP} & 0.3335	\tiny{(±0.02)} & 38.91	\tiny{(±0.55)}      & 3.307	\tiny{(±0.14)} \\
    \midrule
    \multirow{17}[4]{*}{Search} & \textbf{AFP} & 0.6333	\tiny{(±0.01)} & 34.89	\tiny{(±1)}      & 2.941	\tiny{(±0.12)} \\
          & \textbf{AFP-FE} & 0.64	\tiny{(±0.01)} & 36.04	\tiny{(±1)}      & 3.091	\tiny{(±0.12)} \\
          & \textbf{AIFeynman2} & 0.211	\tiny{(±0.02)} & 2240	\tiny{(±250)}      & 3.248	\tiny{(±0.14)} \\
          & \textbf{BSR} & 0.2725	\tiny{(±0.02)} & 22.52	\tiny{(±0.91)}      & 2.92	\tiny{(±0.08)} \\
          & \textbf{DSR} & 0.5625	\tiny{(±0.01)} & 9.465	\tiny{(±0.26)}      & 2.408	\tiny{(±0.07)} \\
          & \textbf{EPLEX} & 0.7372	\tiny{(±0.01)} & 53.14	\tiny{(±0.73)}      & 3.315	\tiny{(±0.11)} \\
          & \textbf{GP-GOMEA} & 0.7381	\tiny{(±0.01)} & 30.27	\tiny{(±0.96)}      & 2.996	\tiny{(±0.08)} \\
          & \textbf{GPlearn} & 0.539	\tiny{(±0.01)} & 19.06	\tiny{(±0.96)}      & 2.151	\tiny{(±0.08)} \\
          & \textbf{Operon} & 0.7945	\tiny{(±0.02)} & 65.69	\tiny{(±1.3)}      & 3.264	\tiny{(±0.06)} \\
          & \textbf{SBP-GP} & 0.7869	\tiny{(±0.01)} & 634	\tiny{(±18)}      & 6.252	\tiny{(±0.22)} \\
          & \textbf{SR4MDL} & 0.6258	\tiny{(±0.01)} & 29.88	\tiny{(±0.56)}      & 2.267	\tiny{(±0.10)} \\
\cmidrule{2-5}          & \textbf{GP} & 0.6881	\tiny{(±0.04)} & 34.84	\tiny{(±2.4)} & 2.452	\tiny{(±0.18)} \\
          & \textbf{EIC-GP} & 0.6433	\tiny{(±0.05)} & 27.48	\tiny{(±2.1)} &  2.202	\tiny{(±0.21)} \\
          &  $\Delta(\%)$     & -6.50\% & -26.00\% & -10\% \\
\cmidrule{2-5}           & \textbf{MCTS} & 0.6038	\tiny{(±0.06)} & 35.91	\tiny{(±1.9)} & 2.463	\tiny{(±0.27)} \\
          & \textbf{EIC-MCTS} & 0.6227	\tiny{(±0.05)} & 33.89	\tiny{(±1.9)} & 1.951	\tiny{(±0.2)} \\
          &  $\Delta(\%)$     & +3.13\% & -5.62\% & -20.80\% \\
    \bottomrule
    \end{tabular}%
  \label{tab:blackbox}%
\end{table*}%

\subsection{Enhance Generative Methods with EIC}
\label{app:generative_results}

We demonstrate that EIC improves the alignment between pre-training formulas and real-world physical formulas by measuring the similarity between randomly generated formulas and EIC-filtered formulas with three datasets of real physical formulas. These datasets include the Feynman dataset, the Strogatz ODE dataset, and the Wiki Named Equation dataset. The first two are derived from the white-box models in SRBench, while the third was collected by \citet{guimera2020bayesian} from Wikipedia, containing over a thousand named formulas. We cleaned and deduplicated the formulas, removing those with special operators (e.g., matrix operations) and inherently redundant formulas (e.g., numerous occurrences of simple expressions like $a\times b$, retaining only one). The final set contains 940 physical formulas.

\begin{figure}[htbp]
    \centering
    \includegraphics[width=0.6\linewidth]{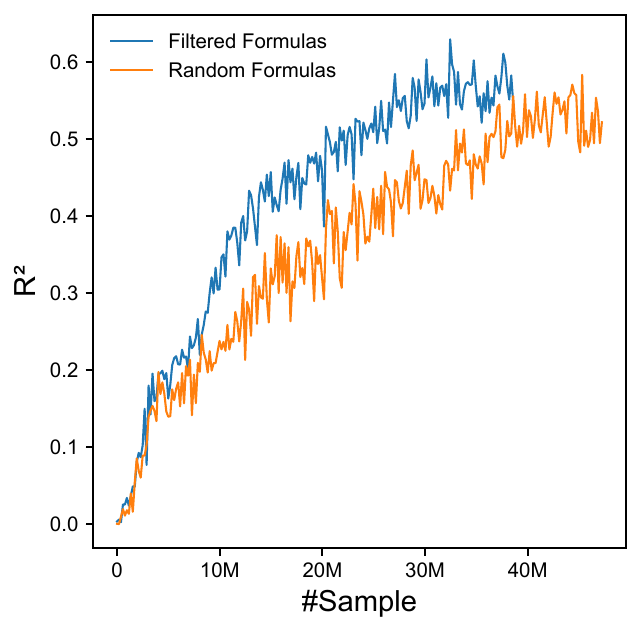}
    \caption{Training process of SNIP using both random and filtered formulas}
    \label{fig:improve_SNIP}
\end{figure}

\begin{figure}[htbp]
    \centering
    \includegraphics[width=\linewidth]{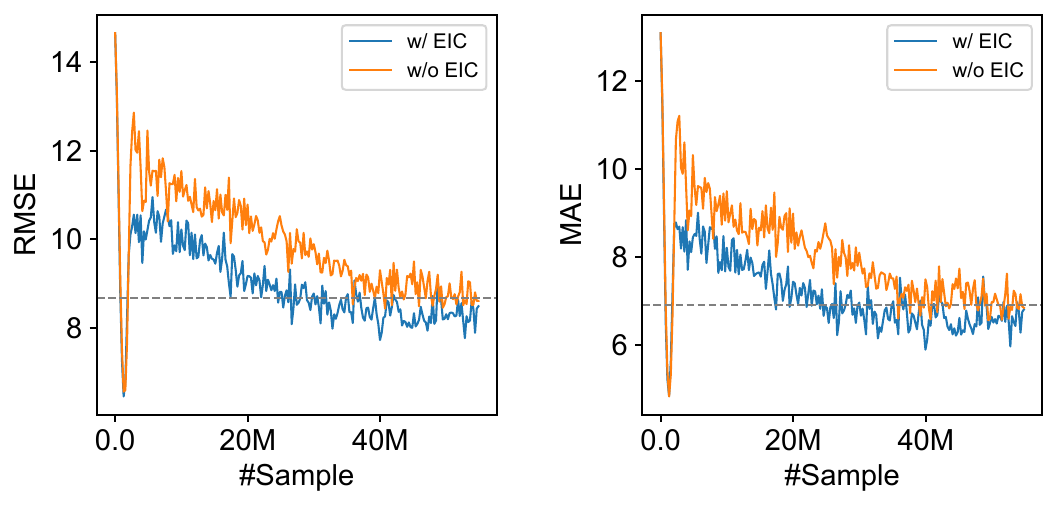}
    \caption{Training process of SR4MDL using both random and filtered formulas}
    \label{fig:improve_SR4MDL}
\end{figure}

In addition to the training process we reported in Figure~\ref{fig:improve_E2ESR}, we also provide the training process of SNIP and SR4MDL in Figure~\ref{fig:improve_SNIP} and Figure~\ref{fig:improve_SR4MDL}. The SNIP model trained on random formulas reaches a final $R^2$ performance of $0.5299$, while it trained on EIC-filtered formulas reaches a final $R^2$ of $0.6016$, which is $13.5\%$ higher. For the SR4MDL, trained on random formulas, it reaches final RMSE and MAE of $8.6778$ and $6.9254$, respectively. While trained on EIC-filtered formulas, it can reach final RMSE and MAE of $8.2319$ and $6.5763$, which are $5.14\%$ and $5.04\%$ higher, respectively.

\subsection{Expert Evaluation Results}
\label{app:expert_rating}

\noindent\textbf{Generation of Formula Pairs for Human Evaluation.}
To rigorously validate the alignment between EIC and human interpretability preferences, we constructed a dataset of paired formulas by strictly controlling for accuracy and complexity. Specifically, we aggregated formulas discovered by search-based symbolic regression methods in Section \ref{sec:results1}, keeping those one- and two-dimensional black-box problems in SRBench (in total of 19 problems). This resulted in a comprehensive candidate pool covering diverse structural patterns. Then, from this aggregated pool, we selected formula pairs to maximize structural divergence (EIC difference) while maintaining strict performance parity. This was formulated as minimizing the following distance function:
\begin{equation}
\begin{aligned}
L(f_{\text{1}}, f_{\text{2}}) = (\text{Complexity}[f_{\text{1}}] - \text{Complexity}[f_{\text{2}}])^2 + \\ (R^2[f_{\text{1}}] - R^2[f_{\text{2}}])^2 - (\text{EIC}[f_{\text{1}}] - \text{EIC}[f_{\text{2}}])^2,
\end{aligned}
\end{equation}
subject to the following constraints to ensure the condition that all else being equal:
\begin{equation}
\begin{aligned}
& |\text{Complexity}[f_{\text{1}}] - \text{Complexity}[f_{\text{2}}]| \leq 2, \\
& |R^2[f_{\text{1}}] - R^2[f_{\text{2}}]| \leq 0.02, \\
& |\text{EIC}[f_{\text{1}}] - \text{EIC}[f_{\text{2}}]| \geq 2, \\
& \max(R^2[f_{\text{1}}], R^2[f_{\text{2}}]) > 0.85.
\end{aligned}
\end{equation}
Here, $f_1$ and $f_2$ represent any two distinct formulas selected from the aggregated pool. The negative sign for the EIC term in $L$ drives the selection toward pairs with the largest possible stability gap, while the constraints ensure that the paired formulas are indistinguishable in terms of fitting accuracy and length. The threshold $R^2>0.85$ ensures that only high-quality candidates are evaluated. This procedure yielded a total of 172 valid formula pairs across the 19 problems.

\noindent\textbf{Human Expert Evaluation.}
For the human rating experiment, we invited 108 volunteer participants with at least a bachelor’s degree in science or engineering. Each participant was randomly assigned 10 formula pairs. For each pair, participants were presented with the data sample visualization, dataset description, and the two candidate formulas alongside their $R^2$ and complexity metrics. They were then asked to choose the more interpretable one based on their domain knowledge and intuition (importantly, EIC scores were withheld to ensure blind evaluation), as demonstrated in Figure~\ref{fig:rating1} and \ref{fig:rating2}. We collected 1080 evaluations in total. After discarding responses completed in under 60 seconds to ensure quality, 840 valid evaluations remained, with each pair assessed on average 4.9 times.

\begin{figure}[htbp]
    \centering
    \includegraphics[width=0.6\linewidth]{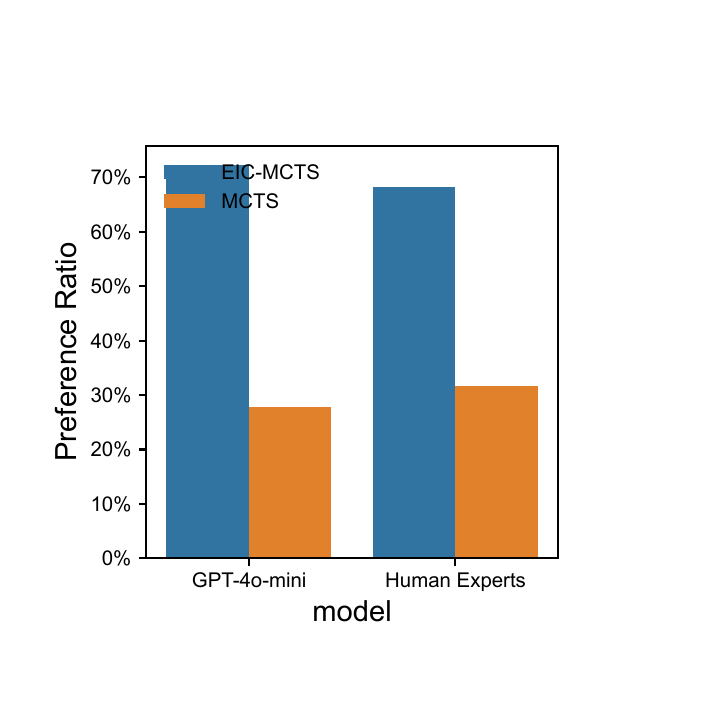}
    \caption{Results of LLM and human experts’ preferences for formula interpretability.}
    \label{fig:LLM_rating2}
\end{figure}

\begin{figure*}[htbp]
    \centering
    \includegraphics[width=\linewidth]{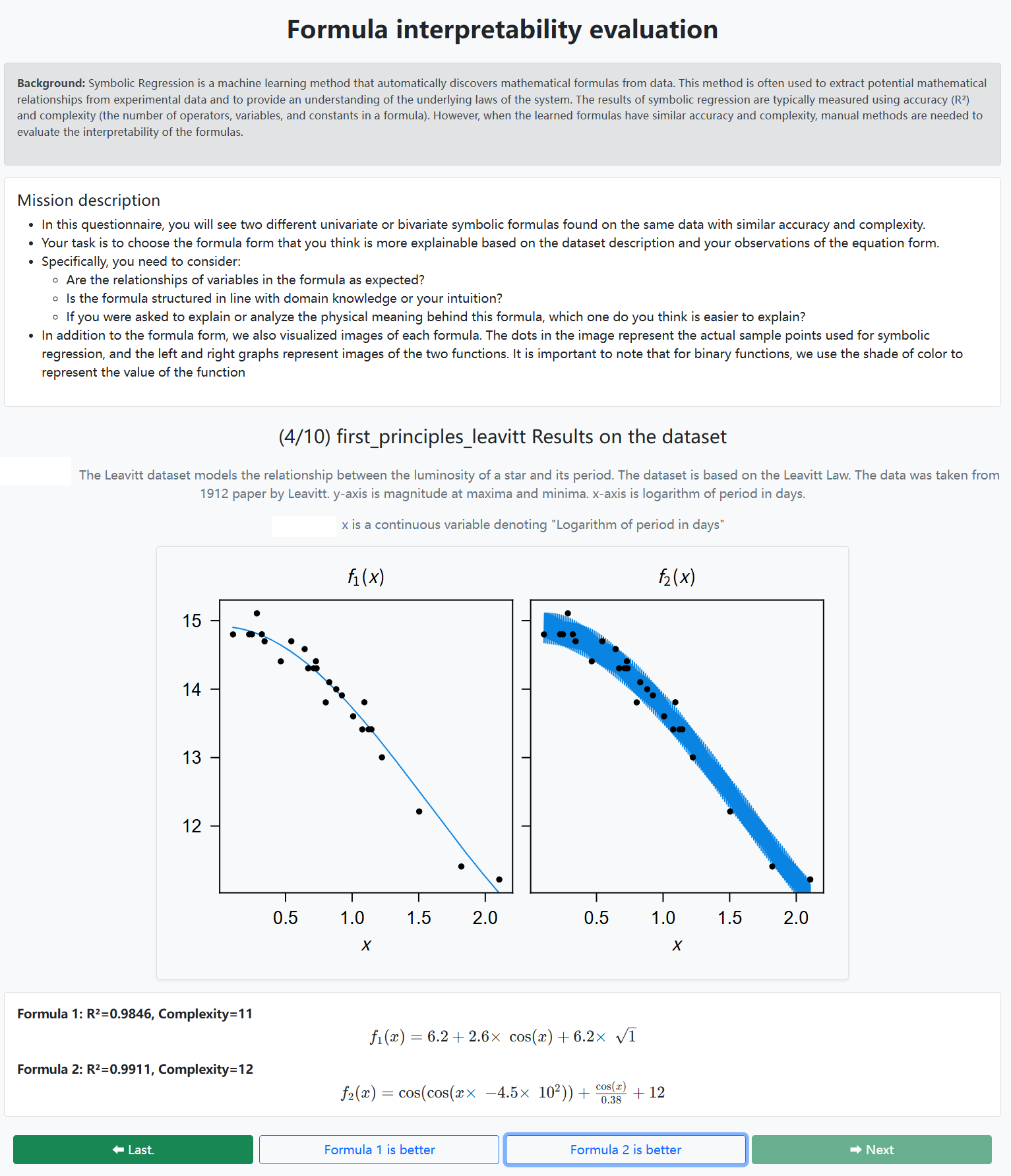}
    \caption{Website interface used by human experts for scoring, where a one-dimensional dataset is demonstrated}
    \label{fig:rating1}
\end{figure*}

\begin{figure*}[htbp]
    \centering
    \includegraphics[width=\linewidth]{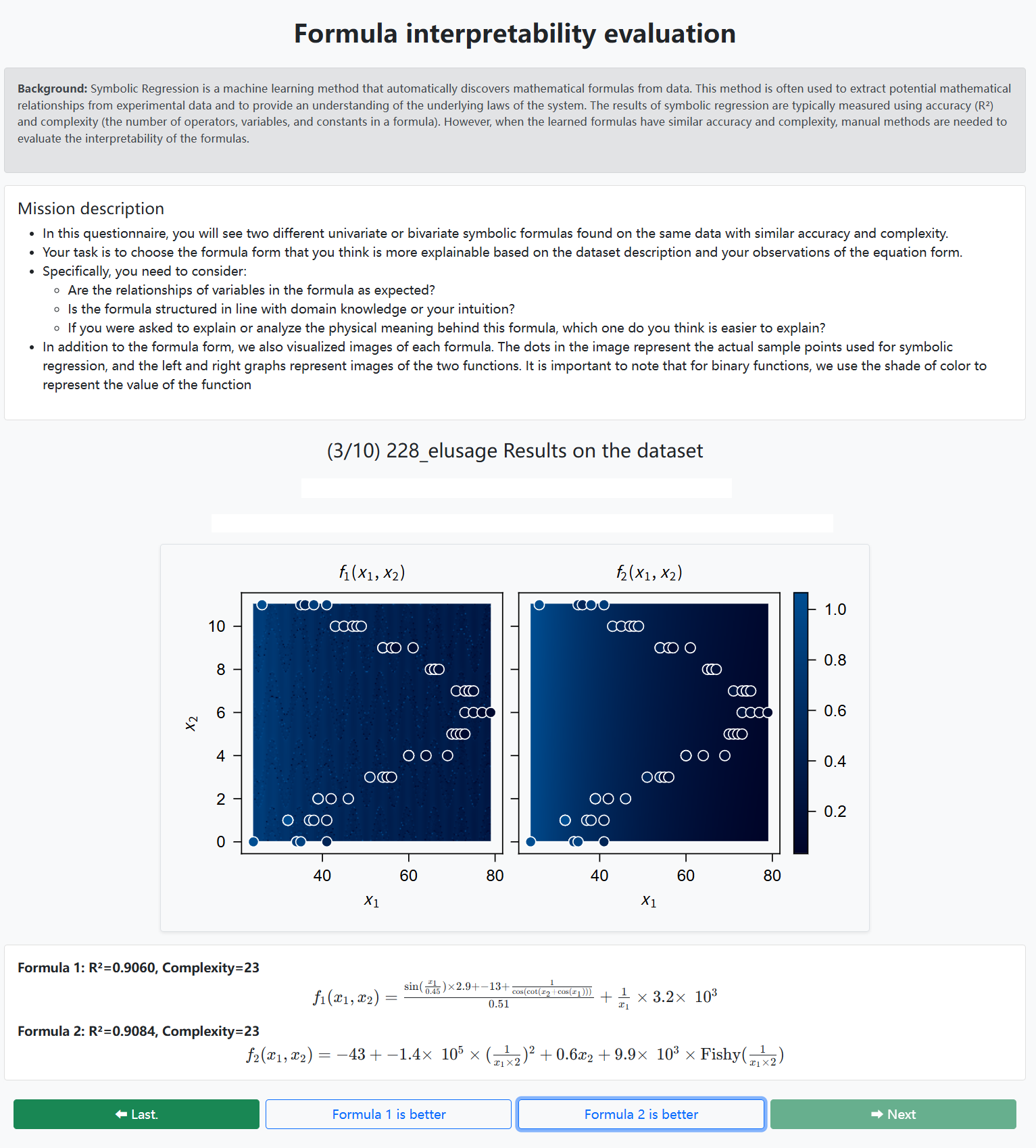}
    \caption{Website interface used by human experts for scoring, where a two-dimensional dataset is demonstrated}
    \label{fig:rating2}
\end{figure*}

\begin{figure*}[htbp]
    \centering
    \includegraphics[width=\linewidth]{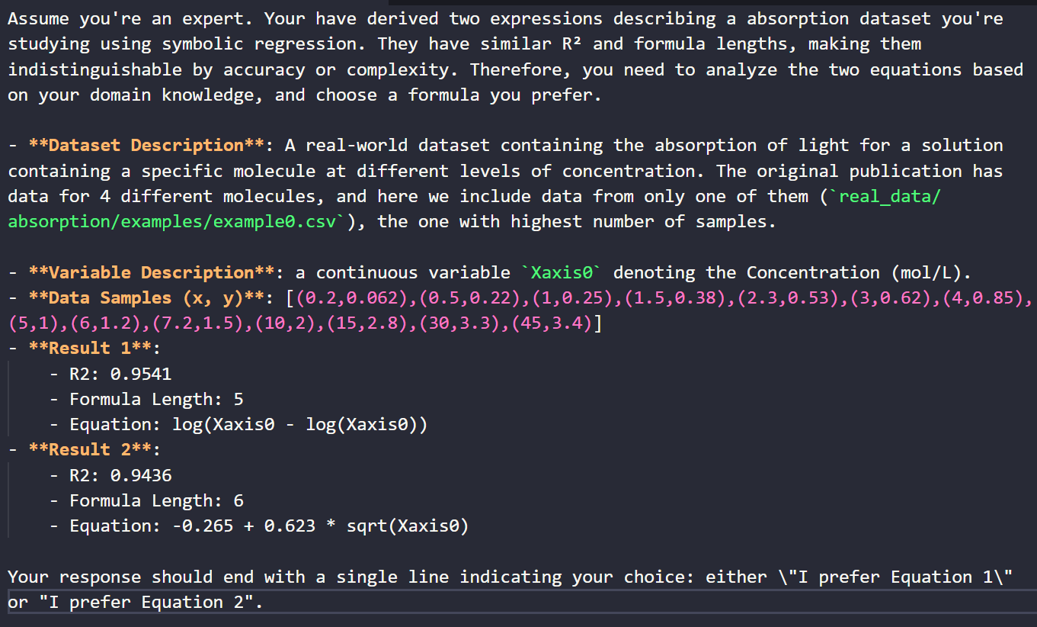}
    \caption{Prompt we used to ask LLM to act as domain experts to evaluate formula interpretability.}
    \label{fig:prompt}
\end{figure*}

\noindent\textbf{LLM Corroboration.}
To provide independent corroboration of the human evaluation results, we also presented each pair to Large Language Models (LLMs) acting as domain experts. We employed two models, GPT-4o-Mini and Qwen3, which are trained on different linguistic distributions, thereby reducing the risk of model-specific bias. We formatted the prompt as shown in Figure~\ref{fig:prompt}, providing the LLMs with the dataset context, sampled points $(x, y)$, mathematical expressions, and performance metrics ($R^2$, Complexity), while withholding EIC scores. We evaluated every formula pair on each LLM five times under a temperature setting of 0.7 to account for sampling variability. The results, shown in Figure~\ref{fig:LLM_rating2}, indicate that LLMs exhibited a preference rate of 72.19\% for the lower-EIC formulas. This alignment with human experts (who preferred lower-EIC formulas in 69.8\% of cases) further validates the reliability of EIC as a proxy for interpretability.

\end{document}